\pdfoutput=1
\documentclass[10pt, logo, onecolumn, copyright]{nv}

\usepackage{amsmath,amsfonts,bm}

\def\eqref#1{equation~\ref{#1}}

\def\1{\bm{1}}

\DeclareMathAlphabet{\mathsfit}{\encodingdefault}{\sfdefault}{m}{sl}
\SetMathAlphabet{\mathsfit}{bold}{\encodingdefault}{\sfdefault}{bx}{n}

\definecolor{linkc}{rgb}{0, 0.44, 0.74}
\definecolor{eqc}{rgb}{1, 0, 0}
\definecolor{newcitecolor}{rgb}{0,0.6,0}
\usepackage{url}
\usepackage{graphicx}
\usepackage{times}
\usepackage{booktabs}
\usepackage{wrapfig}
\usepackage{amsmath}
\usepackage{xspace}
\usepackage{textgreek}
\usepackage{bm}
\usepackage{cleveref}
\crefformat{section}{\S#2#1#3}\Crefformat{section}{\S#2#1#3}
\crefformat{subsection}{\S#2#1#3}\Crefformat{subsection}{\S#2#1#3}
\crefformat{subsubsection}{\S#2#1#3}\Crefformat{subsubsection}{\S#2#1#3}
\crefmultiformat{section}{\S\S#2#1#3}{ and~#2#1#3}{, #2#1#3}{ and~#2#1#3}
\Crefmultiformat{section}{\S\S#2#1#3}{ and~#2#1#3}{, #2#1#3}{ and~#2#1#3}
\crefmultiformat{subsection}{\S\S#2#1#3}{ and~#2#1#3}{, #2#1#3}{ and~#2#1#3}
\Crefmultiformat{subsection}{\S\S#2#1#3}{ and~#2#1#3}{, #2#1#3}{ and~#2#1#3}
\crefrangeformat{section}{\S\S#3#1#4--#5#2#6}\Crefrangeformat{section}{\S\S#3#1#4--#5#2#6}
\usepackage{multirow}
\usepackage[dvipsnames]{xcolor}
\usepackage{colortbl}
\usepackage{orcidlink}
\usepackage{footmisc}
\usepackage{algorithm}
\usepackage{algpseudocode}
\usepackage{amsmath}
\usepackage{amsfonts}
\usepackage{listings}
\usepackage{graphicx}
\usepackage{caption}
\usepackage{float}
\usepackage{enumitem}
\usepackage{csquotes}
\usepackage{marvosym}
\usepackage{subcaption}
\usepackage{placeins}

\definecolor{mygreen}{RGB}{34,139,34}
\definecolor{mylightblue}{RGB}{0,162,230}
\definecolor{deepyellow}{RGB}{255, 215, 0} 
\definecolor{catgray}{gray}{0.92}
\definecolor{nvidiagreen}{HTML}{76B900}

\definecolor{claudepurple}{HTML}{7030A0}

\definecolor{nvgreen}{HTML}{76B900}
\definecolor{mitred}{HTML}{A31F34}

\makeatletter
\def\blfootnote#1{\xdef\@thefnmark{}\@footnotetext{\scriptsize #1}}
\makeatother

\hypersetup{
    colorlinks=true,
    urlcolor=nvidiagreen,
}

\newcommand{\ours}{SANA-Robot\xspace}
\newcommand{\oursbase}{SANA-Robot-base\xspace}
\newcommand{\ourscausal}{SANA-Robot-AR\xspace}
\newcommand{\speedup}[1]{\textcolor{nvidiagreen}{\scriptsize(#1$\times$)}}
\newcommand{\tablegroup}[2]{\rowcolor{catgray}\multicolumn{#1}{c}{\textbf{#2}}}

\definecolor{assistantcolor}{HTML}{76B900}

\usepackage{soul}

\title{SANA-Robot: Efficient World Action Model with Hybrid Memory}

\author{%
\vspace{-1.5em}
Yuyang Zhao\textsuperscript{1*$\dagger$},~
Zekai Li\textsuperscript{1,2*$\dagger$},~
Yicheng Pan\textsuperscript{3$\dagger$},~
Jiaming Tang\textsuperscript{4},~
Zhuoyang Zhang\textsuperscript{4},~
Jiaxu Xing\textsuperscript{1},~

Moritz Reuss\textsuperscript{1},~
Quanquan Peng\textsuperscript{2},~
Junsong Chen\textsuperscript{1,5},~
Enze Xie\textsuperscript{1},~
Zhijian Liu\textsuperscript{2,6},~
Song Han\textsuperscript{1,4}
\par

{\normalsize \textsuperscript{1}NVIDIA ~~
\textsuperscript{2}UCSD ~~
\textsuperscript{3}THU ~~
\textsuperscript{4}MIT ~~
\textsuperscript{5}HKU ~~
\textsuperscript{6}Inco AI}
\par
\vspace{-1mm}
{\normalfont\normalsize \textsuperscript{*}Equal contribution \quad \textsuperscript{$\dagger$}Core contributor} \\
{\normalfont\normalsize Project Page: \href{https://nvlabs.github.io/Sana/Robot/}{\textbf{https://nvlabs.github.io/Sana/Robot}}}
}

\makeatletter
\renewcommand{\maketitle}{\bgroup\setlength{\parindent}{0pt}
	\begin{center}
		{\titlefont \@title\par}%
		\vskip11pt
		{\@author\par}
		\vskip20pt%
	\end{center}
	\egroup
	{{\abscontent}}%
	\thispagestyle{firststyle}
}
\makeatother

\begin{document}

\begin{abstract}
\noindent\textbf{Abstract:}
World action models (WAMs) leverage the world knowledge of video generation models to improve action generation by aligning visual prediction with action prediction.
However, their prediction overhead prevents real-time deployment on edge devices, and they usually lack an effective long-term memory.
We introduce \ours, a 4.5B WAM with minute-level memory that runs in real time on edge devices, built on the following core designs:
(1) hybrid memory at constant cost: our WAM relies on linear attention to maintain a global memory with a fixed-size state, and on sliding window softmax attention layers that propagate previous information, enabling high success rates on tasks requiring whole-episode memory while keeping an efficient and constant inference speed;
(2) a unified model: by dynamically applying different noise levels to action and video tokens, \ours can serve not only as a policy but also as a world simulator;
(3) a full-stack infrastructure for both memory training and inference, including efficient data packing and shuffling as well as memory-phase-based asynchronous execution.
\ours achieves the best performance on the memory dimension of RoboDojo among models of medium size (about 5B) and the highest success rates on LIBERO-Plus, LIBERO-Pro and LIBERO-Max. 
In real-world deployment, \ours takes 135\,ms per action chunk on an RTX 5090 with FP8 GEMMs and 115\,ms with NVFP4 GEMMs, and achieves closed-loop, real-time asynchronous robot control at 30\,Hz on edge devices, demonstrating both the efficiency and the effectiveness of \ours.
\end{abstract}

\begingroup\hbadness=10000
\maketitle\par
\endgroup

\noindent\begin{minipage}{\linewidth}
\centering
\includegraphics[width=1.0\linewidth]{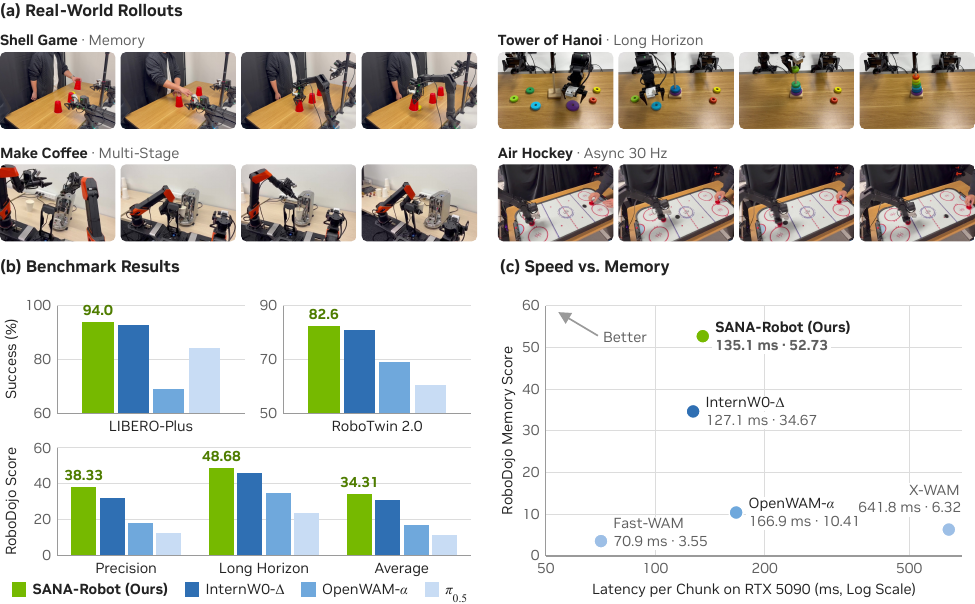}
\captionof{figure}{\ours at a glance. (a) Real-world rollouts. (b) Success rate on LIBERO-Plus and RoboTwin 2.0, and score on the precision and long-horizon dimensions of RoboDojo and its average over the five dimensions, compared with InternW0-$\Delta$, OpenWAM-$\alpha$ and $\pi_{0.5}$. (c) Latency per action chunk against the score on the memory dimension of RoboDojo, all measured on one RTX 5090.}
\label{fig:teaser}
\end{minipage}
\par\vspace{1em}

\section{Introduction}
\label{sec:introduction}

World action models (WAMs) leverage the world knowledge of video generation models~\cite{wan2025wan,agarwal2025cosmos,hacohen2026ltx2} to improve action generation by aligning visual prediction with action prediction~\cite{kim2026cosmospolicy,wang2026openwam,yuan2026fastwam,miao2026internw0delta,guo2026vpp2}.
They now match or surpass vision-language-action models (VLAs)~\cite{black2024pi0,physicalintelligence2025pi05} on manipulation benchmarks~\cite{chen2026robodojo,chen2026robotwin2,fei2026liberoplus}, which makes them a strong candidate for the low-level policy that must run stably and safely on edge devices.
However, current WAMs cannot yet serve as such a policy, for two reasons.
First, video prediction adds a large overhead.
Besides the actions, each action chunk requires denoising the far more numerous video tokens over several steps.
Without optimization, a single request of our 4.5B model takes about 0.6\,s on an RTX 5090, compared with less than 0.1\,s for $\pi_{0.5}$~\cite{physicalintelligence2025pi05}, and a synchronous robot stands still during each request.
Second, many manipulation tasks depend on information that is no longer visible, such as the number of completed steps or the position of an occluded object.
Existing WAMs either act from the current observation alone~\cite{wang2026openwam} or keep only a few sparsely sampled past frames~\cite{miao2026internw0delta,guo2026vpp2}, which miss most of the episode and add to the latency as the memory grows.
Instead, we present \ours, a 4.5B WAM trained from Sana-Video~2~\cite{chen2026sana} that runs in real time on edge devices with a fine-grained, minute-level memory.

\begin{figure}[t]
\centering
\includegraphics[width=\linewidth]{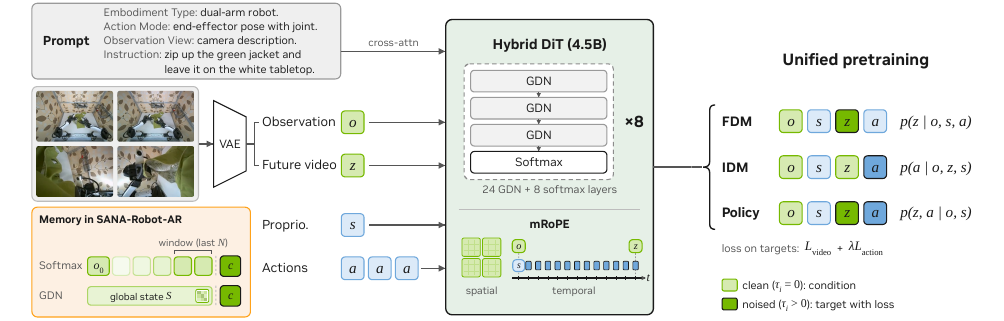}
\caption{Overview of \ours, a unified world action model for FDM, IDM and policy. The video VAE encodes each camera view separately into the observation latent $\bm{o}$ and the future video latent $\bm{z}$, which enter the DiT together with the proprioception $\bm{s}$ and the actions $\bm{a}$; mRoPE places the views on a $2\times2$ grid of positions. The hybrid DiT stacks 8 blocks, each with three GDN layers and one softmax layer.
}
\label{fig:method}
\end{figure}

\ours is a hybrid linear diffusion transformer.
It stacks 8 blocks, each with three Gated DeltaNet (GDN) layers~\cite{yang2024gated} followed by one softmax attention layer.
A GDN layer compresses the past into a fixed-size recurrent state, so its cost does not grow with the context.
A softmax layer attends to every token exactly, at a cost that grows with the context.
This architecture matches the needs of a real-time policy with memory, and we build three core designs on it (\Cref{fig:method}).

\noindent\textbf{Unified World Action Model.}
\ours is pretrained as one model that plays three roles.
As a forward dynamics model (FDM), it predicts the future video caused by given actions and serves as a world simulator for policy evaluation (\Cref{sec:world_simulator}).
As a policy, it predicts the actions and the future video from the current observation and drives manipulation (\Cref{sec:experiments,sec:real}).
As an inverse dynamics model (IDM), it predicts the actions that explain a given video, which makes the video and the actions predicted by the policy more consistent.
All three modes share one token sequence of multi-view video, proprioception and actions, and differ only in which tokens are clean conditions and which are noised targets.
One network thus learns all three roles from the same data, 19.4K hours from 23 datasets of robots, handheld grippers, simulation and human video (\Cref{sec:data}).

\noindent\textbf{Hybrid Memory at Constant Cost.}
Building on its hybrid architecture, we equip \ours with a hybrid memory.
Attention is causal across action chunks, so each chunk reads the past but never the future.
The softmax layers attend to the first frame and to a sliding window over the last few chunks, through which information propagates chunk by chunk.
The GDN layers absorb every past chunk into their fixed-size state and thus keep a memory of the whole episode.
Both memory types cover the long range, at different granularities (\Cref{sec:mem_layers}).
During autoregressive generation, \ours writes each chunk into the memory once and keeps only the GDN state and a rolling KV cache over the sliding window.
Every action chunk therefore has the same cost, while the memory spans up to about one minute on RoboDojo.

\noindent\textbf{Full-Stack Infrastructure.}
The memory changes both training and deployment, and we design the infrastructure of \ours around it.
For training, the memory must be learned on whole episodes, whose lengths vary widely.
We propose multimodal sequence packing with {task-balanced ordering}, which trains on whole episodes without padding and with a stable task mix in every step (\Cref{sec:infra_episode}).
Balanced and shuffled data loading further mixes many datasets in every step (\Cref{sec:infra_loading}).
For deployment, we propose memory RTC for asynchronous execution, which hides the inference latency.
It extends real-time chunking~\cite{black2025rtc} to a policy with memory by keeping one memory per execution phase, so that every request reads a memory laid out as in training (\Cref{sec:inf_pipeline}).
Fused kernels, CUDA graphs, prefix caching and low-bit GEMMs further reduce the latency by $5.3\times$ to 115\,ms per action chunk on an RTX 5090 (\Cref{sec:inf_opt}).
Together, these components enable closed-loop control at 30\,Hz on edge devices.

In summary, \ours is a unified WAM with a hybrid memory that serves as both a policy and an FDM.
It achieves state-of-the-art performance on the memory dimension of RoboDojo~\cite{chen2026robodojo} among models of medium size, and the highest success rates on LIBERO-Plus~\cite{fei2026liberoplus}, LIBERO-Pro~\cite{zhou2025liberopro} and LIBERO-Max~\cite{zhang2026liberomax}.
As a world simulator, it ranks second on WorldArena~\cite{shang2026worldarena} and behaves similarly to the real environment on DROID.
On real-robot tasks, \ours demonstrates precise manipulation, long-term memory and real-time control, and outperforms baselines such as $\pi_{0.5}$~\cite{physicalintelligence2025pi05} and OpenWAM-$\alpha$~\cite{wang2026openwam}.

\FloatBarrier
\section{\ours}
\label{sec:method}

\ours is a world action model (WAM) that jointly generates future video and robot actions with a single diffusion transformer.

\subsection{Unified World Action Model}
\label{sec:method_unified}

\noindent\textbf{{Architecture.}}
We build \ours on Sana-Video~2~\cite{chen2026sana}, a hybrid linear diffusion transformer for video generation.
Its self-attention layers interleave Gated DeltaNet (GDN) linear attention~\cite{yang2024gated} with gated softmax attention, and separate cross-attention layers inject the text.
\ours keeps this single-tower architecture for low latency: the visual and action tokens are placed in one sequence, without a separate action expert.
As shown in \Cref{fig:method}, the sequence holds the multi-view observation latent $\bm{o}$, one proprioception token $\bm{s}$, the future video latent $\bm{z}$ and 32 action tokens $\bm{a}$, where $\bm{s}$ and $\bm{a}$ use the unified action space of \Cref{sec:data_format}.
The causal video VAE of LTX-2~\cite{hacohen2026ltx2} compresses the video by $32\times32\times8$, and $\bm{z}$ encodes the frames of a 32-step chunk at a stride of 4.
The full token sequence is
\begin{equation}
\bm{x} = [\,\bm{o};\ \bm{s};\ \bm{z};\ \bm{a}\,],
\label{eq:unified_sequence}
\end{equation}
{where $\bm{z}$ and $\bm{a}$ attend to $\bm{o}$ and $\bm{s}$ but not the reverse. The text prompt states the embodiment type, the action representation, the camera views and the task instruction (\Cref{sec:data_format}), and every token reads it through the cross-attention layers.}

\noindent\textbf{Positional Encoding.}
We use the 3D multimodal RoPE (mRoPE)~\cite{bai2025qwen3vl} with absolute temporal positions to align the video and action modalities.
A $2\times2$ grid in the RoPE coordinates accommodates the varying camera setups of robot data: one or two head or external cameras occupy the top row, and the wrist cameras the bottom row; empty cells are padded.
Temporally, all tokens share one physical clock: a latent frame is placed at the time of the frames it encodes, the proprioception at the time of the observation, and each action at its own control step.
Video and action tokens that describe the same instant therefore share a temporal position, independent of the frame stride and the control frequency of a dataset.

\noindent\textbf{Unified Training.}
\ours is trained with flow matching~\cite{lipman2023flow,liu2022flow}, and every token $i$ carries its own noise level $\tau_i$ (\Cref{fig:method}).
Clean tokens ($\tau_i=0$) serve as conditions, and noised tokens serve as targets that carry the loss.
The choice of targets thus selects the conditional distribution that one network models:
\begin{equation}
\text{FDM: } p_\theta(\bm{z}\mid\bm{o},\bm{s},\bm{a}),\qquad
\text{IDM: } p_\theta(\bm{a}\mid\bm{o},\bm{z},\bm{s}),\qquad
\text{Policy: } p_\theta(\bm{z},\bm{a}\mid\bm{o},\bm{s}),
\label{eq:modes}
\end{equation}
where every mode is also conditioned on the text prompt.
The Forward Dynamics Model (FDM) simulates the video that an action chunk causes, the Inverse Dynamics Model (IDM) recovers the actions that explain a video, and the Policy predicts both from the current observation and proprioception.
In the Policy mode, the video and the actions of a chunk share one noise level and are denoised jointly.
The loss sums the velocity errors over the target tokens, with the action dimensions that an embodiment lacks masked out (\Cref{sec:data_format}):
\begin{equation}
\mathcal{L} = \mathcal{L}_{\text{video}} + \lambda\,\mathcal{L}_{\text{action}},
\label{eq:loss}
\end{equation}
where we set $\lambda=1$.
The Policy mode drives manipulation (\Cref{sec:experiments,sec:real}), and the FDM serves as a world simulator (\Cref{sec:world_simulator}).

\subsection{\ours Policy}
\label{sec:method_policy}

\noindent\textbf{\oursbase.} \oursbase uses the Policy mode of the unified model: it predicts the action chunk $\bm{a}_c$ and the future video $\bm{z}_c$ from the current observation $\bm{o}_c$ and the proprioception $\bm{s}_c$.
Since the conditions do not attend to the noisy targets (\Cref{eq:unified_sequence}), their keys and values are computed once per request and reused across all denoising steps (\Cref{sec:inf_opt}).

\noindent\textbf{\ourscausal.}\label{sec:method_memory}
A policy that sees only the current observation cannot act correctly when the information that determines the action is occluded, out of view, or carried only by the progress of the task.
\ourscausal therefore conditions the policy on the visual history of the episode.

We treat an episode as a sequence of chunks with chunk-causal attention: chunk $c$ attends to the chunks before it but not to those after it, and within a chunk the video and action tokens attend to each other as in \oursbase.
The memory holds the first frame of the episode, encoded on its own as a sink latent $\bm{o}_0$, and the video latents $\bm{v}_{<c}$ of the past chunks, where $\bm{v}_c$ encodes the frames observed while chunk $c$ is executed; past actions are not stored.
The policy then models
\begin{equation}
p_\theta\big(\bm{z}_c,\bm{a}_c \,\big|\, \bm{o}_0,\ \bm{v}_{<c},\ \bm{o}_c,\ \bm{s}_c\big).
\label{eq:chunk_policy}
\end{equation}

The current observation $\bm{o}_c$ and the proprioception $\bm{s}_c$ read the memory and condition chunk $c$ causally, but they are never written into the memory, so later chunks access the past only through the video latents.
Both choices are necessary (\Cref{sec:mem_obs}): without $\bm{o}_c$, the policy loses precision, and when the targets and $\bm{o}_c$ attend to each other bidirectionally, the policy ignores the memory.

The softmax layers attend to the sink and to the latents of the last $N$ chunks through a sliding window ($N=1$ unless stated otherwise).
Since the cached keys and values of each chunk were computed while it attended to its own window, older history propagates forward chunk by chunk.
The GDN layers compress the whole history into a fixed-size state with a chunk-wise gated delta rule:
\begin{equation}
\bm{S}_c = g_c\,\bm{S}_{c-1} + \bm{K}_c^{\top}\operatorname{diag}(\bm{\beta}_c)\big(\bm{V}_c - \bm{K}_c\,g_c\,\bm{S}_{c-1}\big),
\label{eq:gdn_update}
\end{equation}
where $\bm{S}_c$ is the recurrent state, $g_c$ a data-dependent decay, $\bm{\beta}_c$ the write strengths, and $\bm{K}_c,\bm{V}_c$ the keys and values of chunk $c$.
The queries of chunk $c$ read the decayed previous state together with the keys and values of the chunk itself, so that the tokens within a chunk also see each other:
\begin{equation}
\bm{O}_c = \bm{Q}_c\big(g_c\,\bm{S}_{c-1} + \bm{K}_c^{\top}\operatorname{diag}(\bm{\beta}_c)\,\bm{V}_c\big).
\label{eq:gdn_output}
\end{equation}
Both layer types thus cover the long range at different granularities (\Cref{sec:mem_layers}), and the cost per chunk stays constant.

All tokens keep the physical-time positions of \Cref{sec:method_unified}, except the sink in the softmax layers.
Following StreamingLLM~\cite{xiao2024streamingllm}, we place the sink at most 7 chunks before its reader, so that this distance stays within the range of the 8-chunk training windows.

\noindent\textbf{{Continued Training with Hybrid Memory.}}
\ourscausal is initialized from the unified model and continues pretraining in the Policy mode with the hybrid memory on all data sources, so that the memory is learned before post-training on downstream data.
Since filtering and clip splitting shorten the continuous spans of the data (\Cref{sec:data}), the training windows span up to 8 chunks.
We use chunk-causal teacher forcing: a clean copy of every chunk is written into the memory, while a noised copy, which reads the memory of the chunks before it, carries the loss, so that one forward and backward pass trains all chunks of a window (\Cref{sec:infra_tf}).

\section{Data Pipeline}
\label{sec:data}
Pretraining a world action model across many robots requires data that is both broad and consistent, so that the model learns knowledge of the physical world that is shared across embodiments.
We obtain breadth by collecting 23 datasets of four types: teleoperated robots, handheld UMI devices, simulation and human egocentric video (\Cref{sec:data_sources}).
We obtain consistency by converting them into one format (\Cref{sec:data_format}) and by filtering and balancing them (\Cref{sec:data_filter}).
From 23.8K raw hours, the pipeline produces 19.4K filtered hours and 133M training chunks (\Cref{fig:data}).

\begin{table}[t]
\centering
\caption{Pretraining data before and after filtering. Hours are real time, Views is the number of cameras, and FPS is the training frame-rate tier. Human video is recorded at 30\,Hz and played at 15\,Hz.}
\label{tab:data}
\footnotesize
\setlength{\tabcolsep}{4pt}
\begin{tabular}{llccrrrr}
\toprule
 & & & & \multicolumn{2}{c}{Raw} & \multicolumn{2}{c}{Filtered} \\
\cmidrule(lr){5-6}\cmidrule(lr){7-8}
Dataset & Embodiment & Views & FPS & Hours & Episodes & Hours & Episodes \\
\midrule
\tablegroup{8}{Real Robot} \\
ABC-130K~\cite{allshire2026abc} & YAM & 3--4 & 30 & 3,553.4 & 129,616 & 2,971.3 & 122,559 \\
MolmoAct2-BimanualYAM~\cite{fang2026molmoact2} & YAM & 3 & 30 & 703.8 & 32,234 & 463.3 & 31,958 \\
FineART~\cite{scale2026fineart} & Trossen AI & 4 & 30 & 1,624.1 & 40,088 & 1,405.9 & 39,989 \\
AgiBotWorld-Beta~\cite{agibotworld2025colosseo} & AgiBot G1 & 3 & 30 & 2,706.9 & 165,416 & 2,175.4 & 162,763 \\
AgiBotWorld 2026~\cite{agibot2026world} & AgiBot G2 & 3 & 30 & 591.5 & 28,510 & 453.2 & 27,443 \\
GM-100~\cite{wang2026gm100} & AgiBot G1 / AgileX / R1 Pro & 3 & 15/30 & 567.5 & 45,462 & 499.1 & 45,225 \\
RoboCOIN~\cite{wu2025robocoin} & 8 robots & 3--4 & 30 & 1,023.7 & 128,759 & 789.9 & 123,533 \\
RW-RL~\cite{ye2026evo} & SO-101 / AgiBot G2 / R1 Lite & 3 & 15/30 & 486.0 & 23,161 & 394.5 & 21,985 \\
Galaxea Open-World~\cite{jiang2025galaxea} & Galaxea R1 Lite & 4 & 15 & 527.4 & 22,438 & 430.0 & 22,243 \\
DROID~\cite{khazatsky2024droid} & Franka & 3 & 15 & 346.1 & 57,639 & 317.1 & 57,456 \\
RoboMIND~\cite{wu2024robomind} & Franka & 3--4 & 15 & 42.5 & 17,154 & 34.4 & 16,506 \\
RoboMIND 2.0~\cite{hou2025robomind2} & Franka FR3 (dual) & 4 & 15 & 357.8 & 75,814 & 284.1 & 75,044 \\
RealSource-World~\cite{realsource2025world} & RealSource RS-02 & 3 & 30 & 345.4 & 26,667 & 302.3 & 26,656 \\
PrimeBot~\cite{xu2026primebot} & PrimeBot X2W & 3 & 30 & 502.7 & 30,170 & 456.6 & 30,166 \\
ActionNet~\cite{fourier2025actionnet} & Fourier GR-1 & 1 & 30 & 142.0 & 30,120 & 133.0 & 30,111 \\
Unitree G1 Dex1~\cite{xr-teleoperate} & Unitree G1 & 3 & 30 & 327.7 & 60,066 & 249.1 & 59,507 \\
Unitree WBT~\cite{xr-teleoperate} & Unitree G1 & 3 & 30 & 120.5 & 11,242 & 79.3 & 10,611 \\
Dexora~\cite{zhang2026dexora} & AIRBOT MMK2 & 4 & 20 & 43.2 & 12,045 & 39.9 & 11,932 \\
\textit{Subtotal} & & & & 14,012.3 & 936,601 & 11,478.3 & 915,687 \\
\midrule
\tablegroup{8}{UMI} \\
HyEmbodied-UMI~\cite{zhang2026hyembodied} & UMI gripper & 3 & 30 & 2,161.3 & 250,133 & 2,130.8 & 249,806 \\
\midrule
\tablegroup{8}{Simulation} \\
InternData-A1~\cite{tian2025interndataa1} & Franka / Split ALOHA / Lift2 & 2--3 & 30 & 3,693.0 & 559,256 & 3,091.8 & 557,196 \\
BEHAVIOR-1K~\cite{li2024behavior1k} & Galaxea R1 Pro & 3 & 30 & 1,952.9 & 20,000 & 1,292.3 & 19,996 \\
\textit{Subtotal} & & & & 5,645.9 & 579,256 & 4,384.1 & 577,192 \\
\midrule
\tablegroup{8}{Human Egocentric} \\
EgoDex~\cite{hoque2025egodex} & Human & 1 & 15 & 829.4 & 338,234 & 592.7 & 277,600 \\
EgoVerse~\cite{punamiya2026egoverse} & Human & 1 & 15 & 1,141.4 & 410,172 & 798.4 & 324,076 \\
\textit{Subtotal} & & & & 1,970.8 & 748,406 & 1,391.0 & 601,676 \\
\midrule
Total & & & & 23,790.3 & 2,514,396 & 19,384.2 & 2,344,361 \\
\bottomrule
\end{tabular}
\end{table}

\subsection{Data Sources}
\label{sec:data_sources}
\noindent\textbf{Real Robot Teleoperation.}
Teleoperated demonstrations provide the robot actions that the policy learns to produce.
They make up 14.0K of the raw hours and come from 18 datasets covering 15 robot families.
These include bimanual YAM~\cite{allshire2026abc,fang2026molmoact2}, AgiBot~\cite{agibotworld2025colosseo,agibot2026world}, Trossen~\cite{scale2026fineart}, single- and dual-arm Franka~\cite{khazatsky2024droid,wu2024robomind,hou2025robomind2}, other wheeled, humanoid and arm platforms~\cite{wang2026gm100,wu2025robocoin,ye2026evo,jiang2025galaxea,realsource2025world,xu2026primebot,xr-teleoperate} and dexterous hands~\cite{fourier2025actionnet,zhang2026dexora}.

\noindent\textbf{UMI Recordings.}
Handheld UMI grippers~\cite{chi2024umi} provide end-effector actions without a robot, since the gripper is held by a person and tracked.
HyEmbodied-UMI~\cite{zhang2026hyembodied} contributes 2.2K hours of such recordings.

\noindent\textbf{Simulation.}
Simulation contributes 5.6K raw hours with exact proprioception and action labels.
InternData-A1~\cite{tian2025interndataa1} provides Franka, Split ALOHA and ARX Lift2 scenes, and BEHAVIOR-1K~\cite{li2024behavior1k} provides household tasks for the Galaxea R1 Pro.

\noindent\textbf{Human Egocentric Video.}
Human egocentric video offers visual and task diversity that robot data lack.
EgoDex~\cite{hoque2025egodex} and EgoVerse~\cite{punamiya2026egoverse} contribute 2.0K hours of people manipulating objects in everyday scenes.
Videos recorded with fisheye cameras are undistorted to match the other sources.

\subsection{Unified Format}
\label{sec:data_format}
Mixing these sources requires a common format for video, actions and language.
We therefore assign every dataset one of a few frame-rate tiers, map all actions into one space with a shared physical meaning, and give all clips text prompts of one structure.

\noindent\textbf{Time Base.}
The datasets are recorded at different frame rates and with unsynchronized streams.
We assign each dataset the frame-rate tier (10, 15, 20, 24 or 30\,Hz) closest to the median frame rate of its main camera.
Sources faster than 40\,Hz are downsampled to 30\,Hz; all others keep their original frames.
Within an episode, the frames of the main camera define the time steps.
Every other stream, \textit{i.e.}, the other cameras and the proprioception, commands and gripper of each arm, contributes its latest sample at or before each time step.
Thus no step depends on future measurements, and no value is interpolated.
In every tier, a training chunk covers 32 time steps, with 32 actions and 8 video frames at a stride of 4 after the observation frame.
The physical clock of \Cref{sec:method_unified} accounts for the different durations.
Human egocentric video, recorded at 30\,Hz, is assigned the 15\,Hz tier and therefore plays at half speed, which brings hand motion closer to the speed of teleoperated robots.

\noindent\textbf{Action Space.}
All embodiments share one proprioception and action space, which includes absolute end-effector poses, absolute joint positions, and gripper or dexterous-hand values.
A per-dimension mask removes the slots that an embodiment lacks from both the input and the loss.
{Because the same robot may carry different grippers, for real robots we recompute every end-effector pose at the flange from the joint positions with forward kinematics.}
UMI recordings and human video have no joints and fill only the end-effector and gripper slots; for human video, the wrist poses serve as end-effector poses and a virtual gripper represents each hand.
{Actions are normalized with embodiment-specific statistics, which all datasets of the same robot share.}

\noindent\textbf{Segment and Recaption.}
Several datasets contain episodes longer than 10\,s that span several sub-tasks, and their global instruction alone does not ground each part of the episode in language.
We therefore split such episodes into non-overlapping 10\,s clips (the last one shorter) and recaption each clip with a VLM.
The text prompt combines the embodiment type, the action representation, the view description and the global instruction, and during training adds the step instruction of the clip with probability 0.5, for example:
\definecolor{promptfill}{HTML}{FFFBEA}
\definecolor{promptframe}{HTML}{F1D98A}
\begin{mdframed}[backgroundcolor=promptfill, linecolor=promptframe, linewidth=0.6pt, innertopmargin=4pt, innerbottommargin=5pt, innerleftmargin=6pt, innerrightmargin=6pt, skipabove=5pt, skipbelow=6pt]
\noindent\small
Embodiment Type: dual-arm YAM robot with parallel grippers.\\
Action Mode: robot\_base\_eef; end-effector targets are absolute future poses expressed in the robot base frame, joint targets are absolute future targets, and gripper targets are absolute future targets.\\
Observation View: left wrist-mounted camera.\\
Instruction: Zip up the green jacket and leave it on the white tabletop.\\
Current step: Zip up the green jacket.
\end{mdframed}

\subsection{Filtering and Balancing}
\label{sec:data_filter}
Raw robot data contain defects that the model should not learn from: sensor glitches, idle periods, unusable frames, and video that does not show the recorded motion.
Rather than discarding every episode that contains a defect, our filter removes only the affected time steps and keeps the clean spans in between; it drops an episode entirely only when a defect affects all of it.
Overall, filtering keeps 81.5\% of the raw hours and 93.2\% of the episodes (\Cref{tab:data}).

\noindent\textbf{Kinematic Validity.}
Actions that violate the kinematics of the robot indicate recording errors.
We check every action against the kinematic model of the robot.
We remove actions with invalid joint values, actions whose forward kinematics disagrees with the recorded end-effector pose, and actions whose gripper value lies far outside the range between its 1st and 99th percentiles.

\noindent\textbf{Motion.}
Long idle periods add no supervision, and jumps indicate recording errors.
We remove idle periods in which the joints and the gripper all stay still for more than 0.5\,s, but keep shorter pauses, which are part of how a task is executed.
A step whose speed between consecutive actions exceeds the physical speed limits (2\,rad/s for joints; 0.5\,m/s and 2\,rad/s for the end effector) is a jump; we remove it and split the episode there.

\noindent\textbf{Video Quality.}
Frames that are too dark, overexposed or of very low contrast show little of the scene.
We estimate the brightness and contrast of each frame and remove such frames.

\noindent\textbf{Video-Action Consistency.}
A video that does not show the motion described by the actions pairs every action with a wrong observation.
Frames whose pixels change significantly while the robot is static, and arms whose largest motion is not visible in any of their cameras, are flagged and skipped in pretraining.
We drop an entire episode when its video largely mismatches the proprioception, for example when the video lags behind it.

\begin{figure}[t]
\centering
\includegraphics[width=\linewidth]{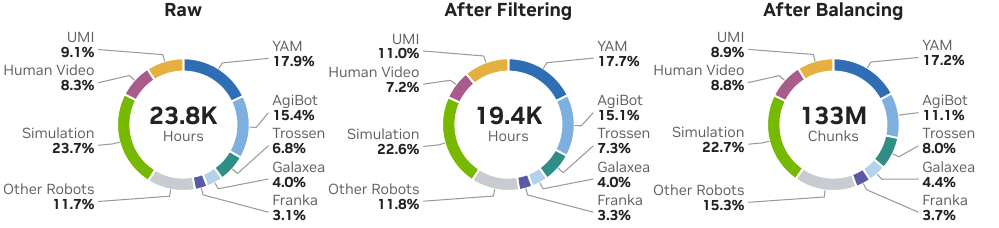}
\caption{Pretraining data by source: shares of the raw hours, of the filtered hours, and of the training chunks per epoch after balancing. Other Robots are the ten smaller robot families of \Cref{tab:data}.}
\label{fig:data}
\end{figure}

\noindent\textbf{Balancing.}
Sampling in proportion to hours would let a few large datasets dominate training, while some high-quality robot families each hold less than 1\%.
We therefore balance the data by embodiment and data quality.
The weight of a robot family grows with its filtered hours raised to the power 0.7, adjusted by a small family-specific prior that reflects data quality.
Within a family, the weight of a dataset grows with its filtered hours raised to the power 0.9.
{Balancing thus gives small, high-quality families a larger share than their hours alone would.}
\Cref{fig:data} shows how filtering and balancing change the composition of the corpus.

\section{Training Infrastructure}
\label{sec:training_infra}
Training throughput determines how much data and how many design variants we can afford to train on.
We raise it at three points: loading the corpus with a balanced mixture (\Cref{sec:infra_loading}), training all chunks of a window in one forward and backward pass (\Cref{sec:infra_tf}), and training \ourscausal on whole episodes of varying length (\Cref{sec:infra_episode}).

\subsection{Data Loading}
\label{sec:infra_loading}

\noindent\textbf{Data Storage.}
We store every dataset as ZIP archives of up to 1,000 episodes each.
An episode keeps its camera videos, its proprioception and action table and its metadata as separate files.
They are stored without compression, so that a data-loader worker reads them directly from the archive.
Episodes are packed in the order of their original keys, so the order of the episodes within an archive follows the source dataset rather than being random.
The loader never expands the corpus into individual samples: it indexes the valid start of every training window in the spans kept by the filters (\Cref{sec:data_filter}) and reads a window only when it is scheduled.

\noindent\textbf{Balanced and Shuffled Sampling.}
{Every window of a dataset should have the same chance of being used, so that each episode contributes in proportion to its length. However, opening archives at random would add RAM and CPU load. We therefore assign each archive to one GPU for the whole epoch, so that only the data-loader workers of that GPU open it.}
The archives are assigned longest first to the GPU with the fewest scheduled windows, which gives all GPUs nearly the same number of windows.
Every GPU then runs the same number of steps.
A GPU forms each batch from windows of one dataset with the same camera layout, resolution and horizon, so that no batch needs padding. Each GPU rotates through a pool of its archives, so that each batch mixes windows from many archives.
Different GPUs draw on different datasets in the same step, so each global batch mixes many datasets.
All GPUs derive the same plan from the random seed and the epoch without communication, which also lets an interrupted run resume at the exact batch where it stopped.

\subsection{Chunk-Causal Teacher Forcing}
\label{sec:infra_tf}
\begin{figure}[t]
\centering
\begin{subfigure}[b]{0.2813\linewidth}
\centering
\includegraphics[width=\linewidth]{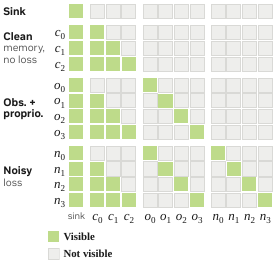}
\caption{Chunk-causal attention mask}
\label{fig:infra_mask}
\end{subfigure}\hfill
\begin{subfigure}[b]{0.3777\linewidth}
\centering
\includegraphics[width=\linewidth]{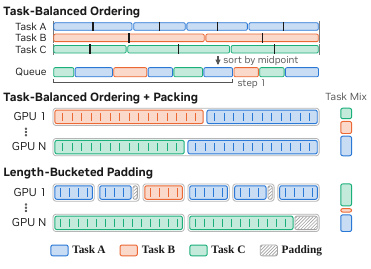}
\caption{Ordering and packing}
\end{subfigure}\hfill
\begin{subfigure}[b]{0.3187\linewidth}
\centering
\includegraphics[width=\linewidth]{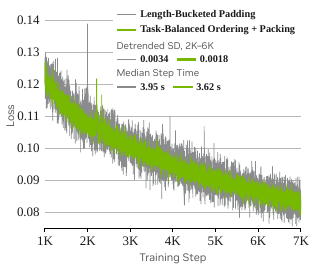}
\caption{Per-step training loss}
\end{subfigure}
\caption{Training \ourscausal on whole episodes. (a) Attention mask of chunk-causal teacher forcing for a window of four chunks: each row is a query and each column a key, and $c_k$, $o_k$ and $n_k$ denote the clean, observation (with proprioception) and noisy entries of chunk $k$. 
(b) Task-balanced ordering and one training step, schematic: the windows of each task are laid end to end on a line from 0 to 1 and sorted by their midpoints into one queue, which packing places on the GPUs without padding; length-bucketed padding wastes compute and skews the task mix of the step. 
(c) Per-step training loss when fine-tuning \ourscausal on RoboDojo: task-balanced ordering with packing halves the detrended loss fluctuation and shortens the median step time.}
\label{fig:infra_packing}
\end{figure}
\ourscausal is autoregressive: it predicts an episode chunk by chunk, and each chunk is conditioned on the memory of the chunks before it (\Cref{eq:chunk_policy}).
Unlike autoregressive video generation, where each frame is conditioned on the history the model has generated itself~\cite{huang2025selfforcing}, robot manipulation is teacher-forced by nature: the memory is written from the observed frames, so the next chunk is always predicted from the real observations.
A naive way to train such a policy samples a segment of an episode and supervises only its last chunk, with the earlier chunks serving as memory, so that each forward pass trains a single chunk.
We instead train every chunk of a window in one forward and backward pass.
Since a diffusion model sees each chunk noised whereas the memory must hold it clean, we place every chunk in the window twice (\Cref{fig:infra_mask}): a clean stream of the video latents builds the memory and carries no loss, and a noisy stream holds, for every chunk $c$, $\bm{o}_c$, $\bm{s}_c$ and the noised targets $\bm{z}_c$ and $\bm{a}_c$, and carries the loss.
The attention keeps this layout chunk-causal: in the softmax layers, noisy chunk $c$ reads the sink, the last $N$ clean chunks before it and its own entries, where $\bm{o}_c$ and $\bm{s}_c$ do not read the targets; in the GDN layers, it reads the state after the sink and the clean chunks before it, and its $\bm{o}_c$ and $\bm{s}_c$ update only a copy of that state, which is discarded after chunk $c$.
Each noisy chunk thus computes the same output as the forward pass that predicts it at deployment, and one pass over a sequence about twice as long as the window trains all $L$ chunks of the window, which the naive scheme trains in $L$ passes over segments of $1$ to $L$ chunks.

\subsection{Variable-Length Causal Training}
\label{sec:infra_episode}
For every downstream dataset, we fine-tune \ourscausal on whole episodes, so that its memory spans the full task as it does at deployment.
Each training window is then one complete episode of a single task, and its length varies with the task and the demonstration.

In causal training, the windows vary in length, which raises three problems in turn:
1) GPUs with shorter windows wait for the one with the longest at every synchronization, which lowers the model FLOPs utilization (MFU);
2) padding the windows to a common length, the standard remedy, equalizes the GPUs but spends compute on the padding;
3) grouping windows of similar length to reduce the padding unbalances the tasks: windows of similar length often come from the same task, so the task mix of a step deviates strongly from that of the data and the loss fluctuates from step to step.
We therefore order and pack the windows so that every step balances the load across GPUs, spends no compute on padding and keeps the task mix of the data.

\noindent\textbf{Task-Balanced Ordering.}
We first arrange the windows of an epoch into one queue in which every stretch holds each task near its share of the data (\Cref{fig:infra_packing}b).
For each task, we shuffle its windows and lay them end to end on a line from 0 to 1, each window spanning a length in proportion to its number of chunks.
Each window is keyed by the midpoint of its span, and the windows of all tasks are sorted by this key.
Since all tasks advance along their lines at the same rate, up to any key $x$ every task has placed a fraction $x$ of its chunks, to within half a window, so every stretch of the queue keeps the task mix of the data, while the order within each task stays random.
The packing below considers only the window lengths and the queue order, so every step inherits this balanced mix.
On RoboDojo, even the least balanced steps then come close to the task mix of the whole epoch.
Compared with length-bucketed padding, the detrended standard deviation of the per-step loss drops from 0.0034 to 0.0018 over steps 2K--6K (\Cref{fig:infra_packing}c).

\noindent\textbf{Sequence Packing for Load Balancing.}
Packing aims to give every GPU the same load without spending compute on padding, so that no GPU waits for another when they synchronize and the MFU stays high.
Following the queue, we fill each GPU with windows of different lengths up to a fixed budget of chunks per step, as in sequence packing for LLM training~\cite{krell2022packing}.
Following the implementation of the variable-length attention of FlashAttention~\cite{dao2023flashattention2}, the windows packed on a GPU do not interact: attention, the GDN state and the RoPE positions all restart at every window.
On RoboDojo, packing cuts the median step time from 3.95\,s to 3.62\,s ($8\%$), although each step carries slightly more chunks (\Cref{fig:infra_packing}c).

\subsection{Robot Pretraining}
\label{sec:pretrain}

\noindent{\textbf{Implementation Details.} We initialize \ours from Sana-Video~2~\cite{chen2026sana}, whose backbone interleaves GDN and softmax attention, with softmax attention in a quarter of the self-attention layers.
Each sample is a chunk of $C=32$ control steps: the observation latent of the current frame, the proprioception, 32 actions, and one future latent that encodes the frames for the next 32 actions.
The camera views of every dataset are placed on the $2\times2$ RoPE grid at a base resolution of 256 pixels, and the data loader rotates each GPU through a pool of 32 archives.
We train with AdamW and a cosine learning-rate schedule, clip the gradient norm at 1.0, use BF16 mixed precision with activation checkpointing, and keep an exponential moving average (EMA) of the weights.
Flow matching uses a timestep shift~\cite{esser2024sd3} of 5 for the video and 1 for the actions.}

\noindent{\textbf{Base Training.} We first train the unified model of \ours on single chunks for 100k iterations with a peak learning rate of $10^{-4}$, assigning each sample to the Policy, FDM or IDM mode at a ratio of 8:1:1. The base model is trained with a global batch size of 4,096 on 64 GB300 GPUs. For classifier-free guidance, we randomly drop the task instruction with probability 0.1.}

\noindent{\textbf{Memory Training.} We then continue training the Policy with the hybrid memory for 50k iterations with a peak learning rate of $5 \times 10^{-5}$ on windows of up to 8 chunks, using the chunk-causal teacher forcing of \Cref{sec:infra_tf}.}

\section{Real-Time Inference}
\label{sec:inference}

Real-time closed-loop control requires the robot to run stably at its control frequency, 30\,Hz in our deployment, without any idle time.
This places two requirements on inference.
1) The pipeline must be asynchronous, since in synchronous execution the robot idles during every request (\Cref{sec:inf_pipeline}).
2) Inference must be fast, since a lower latency allows a smaller asynchronous delay, so that each chunk is planned from a more recent observation and asynchronous execution loses less accuracy (\Cref{sec:inf_opt}).
\Cref{sec:inf_latency} reports the resulting latency on server and edge GPUs.

\begin{figure}[t]
\centering
\includegraphics[width=\linewidth]{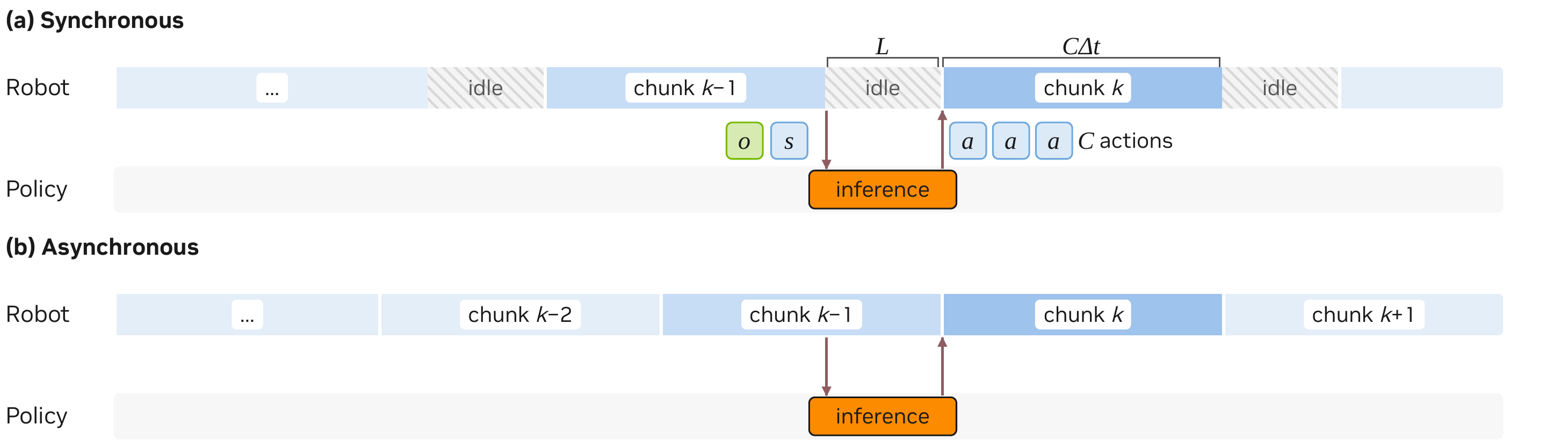}
\caption{Synchronous and asynchronous inference.
(a) Synchronous execution: the robot idles for the latency $L$ of each request and then executes all $C$ actions of the new chunk.
(b) Asynchronous execution: the request at $t_k$ is issued when $d$ actions of the current chunk remain and runs while the robot executes them; the first $d$ actions of chunk $k$ are dropped, and the robot executes the next $P=C-d$.
}
\label{fig:inf_pipeline}
\end{figure}

\begin{figure}[t]
\centering
\includegraphics[width=\linewidth]{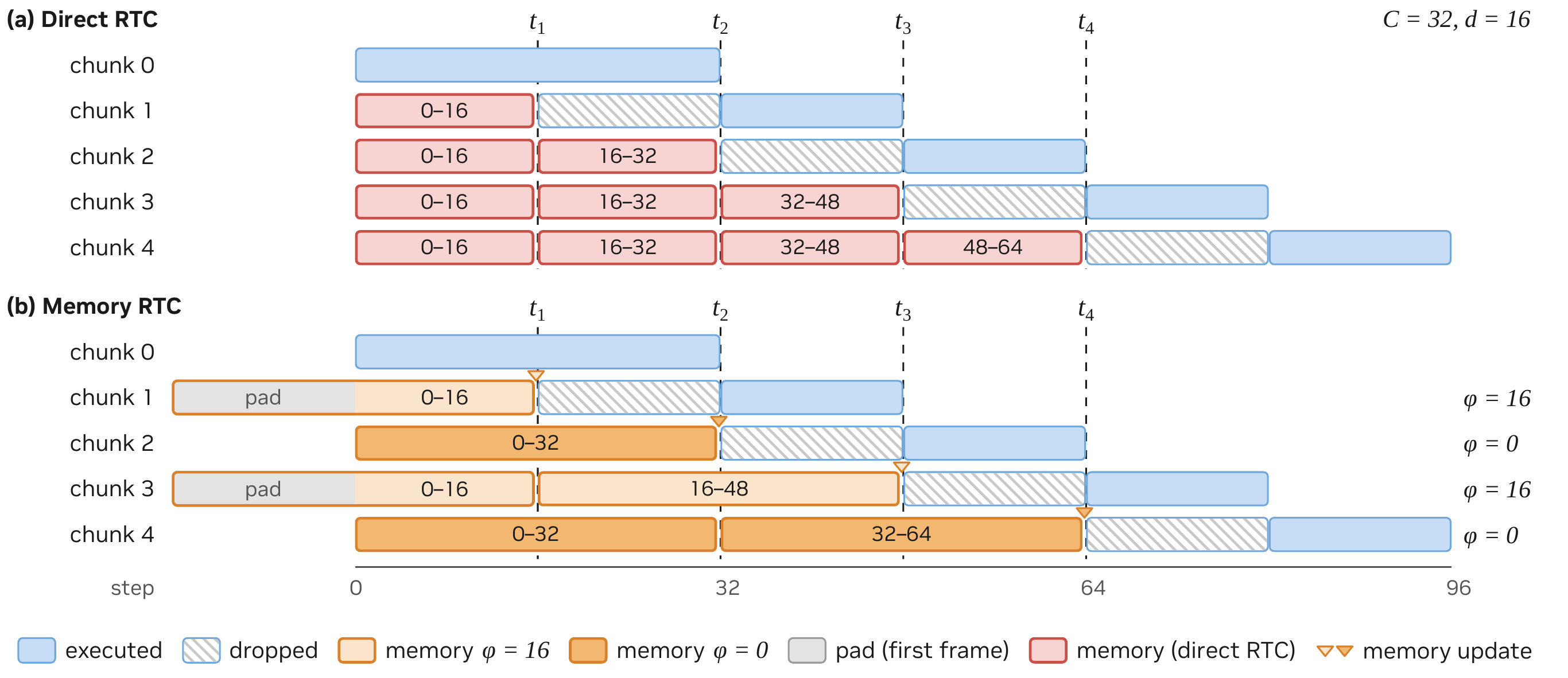}
\caption{
Asynchronous execution of \ourscausal with $C=32$ and $d=16$. Request $k$ starts at step $t_k$, drops the first $d$ actions of its chunk and executes the next $C-d$. (a) Direct RTC keeps a single memory and writes the $C-d$ executed steps as one memory chunk, so its memory chunks are shorter than in training. (b) Memory RTC keeps one memory per phase $\varphi$, whose chunks span $C$ steps as in training; at each $t_k$ (triangles), the window that has just ended is written into the memory of phase $\varphi_k$, here every 16 steps, alternating between the two memories.}
\label{fig:inf_phase}
\end{figure}

\subsection{Deployment Pipeline}
\label{sec:inf_pipeline}

In synchronous execution, the policy is requested once per chunk, and the robot executes all $C$ actions of the new chunk before the next request; with a control period $\Delta t$ and a per-request latency $L$, each chunk takes $C\Delta t + L$, of which the robot idles for $L$.
Asynchronous execution~\cite{black2025rtc,li2026flashvla} removes this idle time by overlapping inference with execution: the request for chunk $k$ is submitted at step $t_k=k(C-d)$, when $d$ actions of the current chunk remain to be executed. As long as $L \le d\,\Delta t$, there will be no idle time with the first $d$ actions of chunk $k$ dropped, and the robot executes the next $C-d$ actions (\Cref{fig:inf_pipeline}b).

\noindent\textbf{\ourscausal in Synchronous Execution.}
The softmax layers maintain a sliding-window KV cache over the sink and the last $N$ chunks, and the GDN layers maintain a global KV state $\bm{S}$ over all past chunks.
At the start of an episode, the first frame is encoded as the sink latent and prefilled into both.
Each request then takes three steps:
(i) \emph{encoding}: the VAE encodes the frames observed during the last $C$ steps into a video latent and the current frames into $\bm{o}$;
(ii) \emph{prefill}: the video latent is forwarded once and written into the memory, updating the sliding-window KV cache in a FIFO manner and the GDN state by \cref{eq:gdn_update}, and $\bm{o}$ and $\bm{s}$ are forwarded conditioned on the memory, without being written into it;
(iii) \emph{generation}: $\bm{z}$ and $\bm{a}$ are denoised in $K$ steps, conditioned on the memory and the prefilled conditions.
Although \ourscausal uses $N=1$ by default, the cost of a request is nearly constant in $N$ rather than linear in it.
Only the 8 softmax layers read the window, from a KV cache filled once per chunk, while the 24 GDN layers keep a state of constant size; from $N=1$ to $N=4$, the latency on RTX 5090 rises by only 2.5\,ms in BF16, from 173.3 to 175.8\,ms.

\noindent\textbf{\ourscausal in Asynchronous Execution (Memory RTC).}
In training, a chunk starting at step $t$ reads the memory:
\begin{equation}
\mathcal{M}(t) = \{\bm{o}_0,\ \bm{v}(t),\ \bm{v}(t-C),\ \bm{v}(t-2C),\ \dots\},
\label{eq:phase_memory}
\end{equation}
where $\bm{v}(\tau)$ encodes the frames observed in the $C$ steps before step $\tau$, with copies of the first frame before the start of the episode.
In asynchronous execution, however, consecutive requests are only $C-d$ steps apart, so a single memory can serve all requests only by breaking the chunk structure of training: with $C=32$ and $d=16$, every memory chunk covers 16 steps instead of 32 (\Cref{fig:inf_phase}a), which raises the action error (\Cref{fig:inf_async_error}).

Therefore, we propose \textbf{memory RTC}, extending real-time chunking (RTC)~\cite{black2025rtc} to policies with a chunk-causal memory. The chunk boundaries $t, t-C, t-2C, \dots$ of $\mathcal{M}(t)$ share the remainder $\varphi = t \bmod C$, which we call the phase of the memory.
In asynchronous execution, the phase $\varphi_k$ of request $k$ changes with $t_k$.
Since $t_k=k(C-d)$, $\varphi_k$ cycles through $C/\gcd(C,d)$ phases, e.g., two phases for $d=16$ and four for $d=8$.
We keep $d$ a multiple of the frame stride (\Cref{sec:data_format}), so that every phase falls on an encoded frame.
Memory RTC therefore maintains one memory per phase (\Cref{fig:inf_phase}b), and each request reads the memory of its own phase, which equals $\mathcal{M}(t_k)$ in \cref{eq:phase_memory}.
This is possible because training starts windows at every phase of the chunk grid, so every phase memory matches a memory seen in training.
While the robot runs, $\bm{v}(t)$ is written into the memory of phase $\varphi$ at every step $t$ with $t \bmod C = \varphi$ (triangles in \Cref{fig:inf_phase}b).
Memory RTC thus trades one copy of the memory per phase for exact training-time memories, without recomputing the history at any request.
We evaluate it on real-robot tasks in \Cref{sec:real_tasks}.

\subsection{Inference Optimization}
\label{sec:inf_opt}

\begin{figure}[t]
\centering
\begin{subfigure}[b]{0.284\linewidth}
\centering
\includegraphics[width=\linewidth]{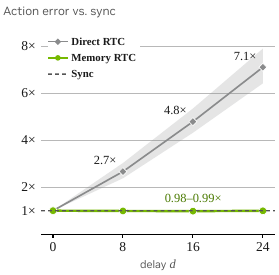}
\caption{Asynchronous action error}
\label{fig:inf_async_error}
\end{subfigure}\hspace{1.5em}
\begin{minipage}[b]{0.502\linewidth}
\centering
\begin{subfigure}[b]{\linewidth}
\centering
\includegraphics[width=\linewidth]{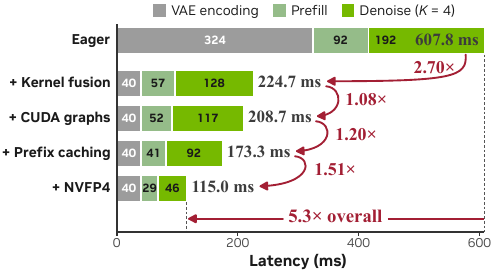}
\caption{Latency of \ourscausal ($N=1$)}
\end{subfigure}
\end{minipage}
\caption{(a) Open-loop end-effector position error of asynchronous execution relative to synchronous execution ($C=32$) on 70 RoboDojo episodes.
Memory RTC stays at the error of synchronous execution for every $d$, whereas direct RTC
degrades as $d$ grows.
(b) Per-request latency of \ourscausal on RTX 5090 as the optimizations are added cumulatively.
}
\label{fig:inf_opt}
\end{figure}

A request runs $K$ denoising steps of the 4.5B-parameter backbone, plus the prefill, which for \ourscausal also writes the last chunk into the memory.
In eager PyTorch, a large part of its latency lies outside the GEMMs: in the GDN layers, whose computation runs as a chain of small kernels, in the normalization, rotary and modulation operations around each GEMM, in the video VAE, and in recomputing the conditions in every forward pass.
We reduce it with kernel fusion, CUDA graphs, prefix caching and, optionally, low-bit GEMMs; \Cref{fig:inf_opt}b shows the latency of \ourscausal as they are added one by one.

\noindent\textbf{Kernel Fusion.}
We implement fused Triton kernels for the operations outside the GEMMs:
1) each GDN layer runs as a single kernel, and for \ourscausal one kernel updates the GDN state when a chunk is written into the memory while another reads it;
2) RMSNorm and RoPE of query and key are fused into one kernel;
3) the attention residuals~\cite{kimi2026attnres}, the adaLN modulation and the gated residual connections around each sublayer are fused into one kernel;
4) the SwiGLU activation is fused into the epilogue of the gate and up projections.
The kernels are tuned separately for each GPU, and all of them match eager execution bitwise or within BF16 rounding.
The VAE encoder is compiled as well.

\noindent\textbf{CUDA Graphs.}
We capture each request of \ourscausal as two CUDA graphs, one writing the last chunk into the memory and one prefilling the conditions and generating, both on the memory in static buffers.
Under asynchronous execution, the memory of each phase is copied into these buffers before its request, so one set of graphs serves all phases.
The compiled VAE encoder is captured as well.

\noindent\textbf{Prefix Caching.}
The conditions, including text prompt and observation, do not attend to the targets (\Cref{sec:method_policy}), so their keys and values are the same in all $K$ denoising steps.
Therefore, we pre-compute them before the denoising steps and reuse them during multiple denoising steps.

\noindent\textbf{Low-Bit GEMMs.}
On edge devices, the GEMM throughput is the main bottleneck once the remaining operations are fused.
We therefore run the large linear layers of the backbone in FP8 (E4M3) with per-tensor scaling, or in NVFP4 (E2M1 with an FP8 scale per block of 16 elements), both with FP32 accumulation, which further reduces latency with only a marginal loss in accuracy (\Cref{sec:real_tasks}). 

\subsection{Latency and On-Edge Deployment}
\label{sec:inf_latency}

\Cref{tab:latency} reports the per-request latency of \ourscausal on RTX PRO 6000, RTX 5090 and Jetson Thor, with $K=4$ denoising steps and three camera views.
We measure the full request as deployed in synchronous execution, i.e., the sum of the latencies of all its components, which run one after another: video encoding, prefill (including the memory) and $K$ denoising steps. The text prompt and the sink, encoded once per episode, are excluded.
\Cref{fig:inf_opt}b breaks this latency down into its stages on RTX 5090.

\begin{table}[t]
\centering
\caption{Per-request latency (ms) of \ourscausal ($N=1$) in synchronous execution with $K=4$: the sum of video encoding, prefill and $K$ denoising steps. Opt.: kernel fusion, CUDA graphs and prefix caching in BF16; +FP8 and +NVFP4: with low-bit GEMMs. Speed-ups over eager in parentheses.}
\label{tab:latency}
\small
\begin{tabular}{lcccc}
\toprule
GPU & Eager & Opt. & +FP8 & +NVFP4 \\
\midrule
RTX PRO 6000 & 573.3 & 175.0 \speedup{3.3} & 140.6 \speedup{4.1} & 123.1 \speedup{4.7} \\
RTX 5090 & 607.8 & 173.3 \speedup{3.5} & 135.1 \speedup{4.5} & 115.0 \speedup{5.3} \\
Jetson Thor & 1390.1 & 420.2 \speedup{3.3} & 333.6 \speedup{4.2} & 288.0 \speedup{4.8} \\
\bottomrule
\end{tabular}
\end{table}

Combining the asynchronous pipeline of \Cref{sec:inf_pipeline} with the optimizations of \Cref{sec:inf_opt}, \ourscausal runs in real time on all three devices.
The optimizations reduce its latency by $3.3$--$3.5\times$ in BF16, $4.1$--$4.5\times$ with FP8 GEMMs and $4.7$--$5.3\times$ with NVFP4 GEMMs, to 115\,ms per request on RTX 5090 and 288\,ms on Jetson Thor.
Under the real-time condition $L\le d\,\Delta t$, a 30\,Hz control loop is thus met with $d=8$ on RTX PRO 6000 and RTX 5090 ($d=4$ with NVFP4) and with $d=16$ on Jetson Thor ($d=12$ with FP8 or NVFP4).
At these delays, memory RTC keeps the action error of synchronous execution (\Cref{fig:inf_async_error}).
We demonstrate 30\,Hz closed-loop control on the dynamic Air Hockey task in \Cref{sec:real_tasks}.

\section{Simulation Benchmark Results}
\label{sec:experiments}
On each benchmark, we post-train \ourscausal from the pretrained model on whole episodes of the demonstrations, which fill every GPU up to a fixed budget of chunks (\Cref{sec:infra_episode}).
All runs use AdamW with 1,000 warmup iterations and a cosine schedule that decays to 1\% of the peak learning rate, and each run stops before the end of its schedule.
Baseline results are taken from the official leaderboards or from the papers that report them.

\subsection{RoboDojo}
\label{sec:sim_robodojo}

\begin{table}[t]
\centering
\caption{Results on RoboDojo~\cite{chen2026robodojo} (simulation). Each cell: average score / success rate (\%). Generalization is the mean of the standard and randomized settings, and the average is taken over the five dimensions. All results, including \ourscausal, are from the official leaderboard.}%
\label{tab:sim_robodojo}
\small
\setlength{\tabcolsep}{3.5pt}
\begin{tabular}{lcccccc}
\toprule
Method & Generalization & Precision & Long horizon & Memory & Open & Avg. \\
\midrule
\tablegroup{7}{VLA} \\
$\pi_0$~\cite{black2024pi0} & 3.95 / 2.56 & 3.56 / 0.75 & 6.19 / 2.00 & 3.47 / 2.11 & 0.25 / 0.25 & 3.48 / 1.53 \\
StarVLA-$\alpha$~\cite{ye2026starvlaalpha} & 3.94 / 2.34 & 9.90 / 4.33 & 14.15 / 6.50 & 3.34 / 2.44 & 0.67 / 0.58 & 6.40 / 3.24 \\
X-VLA~\cite{zheng2025xvla} & 10.47 / 6.78 & 18.32 / 12.00 & 16.53 / 9.75 & 4.76 / 3.56 & 0.55 / 0.50 & 10.13 / 6.52 \\
$\pi_{0.5}$~\cite{physicalintelligence2025pi05} & 13.38 / 8.17 & 12.40 / 5.50 & 23.54 / 14.67 & 5.89 / 4.67 & 1.98 / 1.67 & 11.44 / 6.93 \\
Spatial Forcing~\cite{li2025spatialforcing} & 14.12 / 9.34 & 17.32 / 10.58 & 23.26 / 14.58 & 5.43 / 4.11 & 1.78 / 1.58 & 12.38 / 8.04 \\
Hy-Embodied-0.5-VLA~\cite{zhang2026hyembodied} & 11.78 / 8.39 & 13.81 / 8.00 & 25.74 / 14.92 & 13.37 / 12.11 & 0.65 / 0.58 & 13.07 / 8.80 \\
Xiaomi-Robotics-1~\cite{xiaomi2026robotics1} & 23.55 / 17.00 & 26.69 / 18.83 & 38.39 / 23.67 & 7.81 / 6.56 & 3.94 / 3.58 & 20.07 / 13.93 \\
Galaxea G0.5~\cite{liu2026g05} & 18.46 / 12.83 & 28.25 / 20.42 & 44.12 / \textbf{32.25} & 8.61 / 7.33 & 1.73 / 1.58 & 20.23 / 14.88 \\
DM0.5~\cite{dexmal2026dm05} & 15.78 / 10.95 & 24.82 / 16.75 & 33.70 / 19.50 & 47.74 / 47.44 & 2.43 / 2.08 & 24.90 / 19.34 \\
\midrule
\tablegroup{7}{WAM} \\
Fast-WAM~\cite{yuan2026fastwam} & 2.34 / 1.11 & 1.96 / 0.00 & 9.14 / 5.17 & 3.55 / 3.44 & 0.42 / 0.42 & 3.48 / 2.03 \\
AHA-WAM~\cite{cai2026ahawam} & 5.79 / 3.28 & 5.86 / 2.42 & 8.61 / 2.67 & 2.97 / 2.78 & 0.88 / 0.83 & 4.82 / 2.39 \\
GigaWorld-Policy~\cite{ni2026gigaworld} & 5.35 / 2.89 & 6.15 / 1.83 & 15.51 / 8.92 & 3.46 / 2.22 & 0.54 / 0.50 & 6.20 / 3.27 \\
X-WAM~\cite{guo2026xwam} & 7.39 / 3.33 & 6.72 / 1.83 & 17.47 / 9.08 & 6.32 / 4.67 & 0.57 / 0.25 & 7.69 / 3.83 \\
OpenWAM-$\alpha$~\cite{wang2026openwam} & 20.71 / 14.84 & 18.45 / 9.25 & 34.93 / 25.33 & 10.41 / 9.11 & 1.41 / 1.08 & 17.18 / 11.92 \\
InternW0-$\Delta$~\cite{miao2026internw0delta} & 30.09 / 22.78 & 31.98 / 23.25 & 46.29 / 29.33 & 34.67 / 34.00 & 10.84 / 10.17 & 30.77 / 23.91 \\
\midrule
\tablegroup{7}{LLM as policy} \\
GPT-6 Astra~\cite{zhang2026gpt6astra} & \textbf{33.36} / \textbf{30.50} & 12.65 / 4.00 & 21.45 / 8.25 & 43.04 / 38.67 & \textbf{34.36 / 31.00} & 28.97 / 22.48 \\
\midrule
\textbf{\ourscausal} & 26.61 / 21.11 & \textbf{38.33} / \textbf{28.67} & \textbf{48.68} / 28.75 & \textbf{52.73} / \textbf{52.33} & 5.21 / 4.58 & \textbf{34.31} / \textbf{27.09} \\
\bottomrule
\end{tabular}
\end{table}

{The benchmark~\cite{chen2026robodojo} contains 42 bimanual simulation tasks on the ARX X5 robot, grouped into five capability dimensions: generalization; precision; long-horizon execution; memory, whose tasks depend on information observed earlier in the episode; and open-vocabulary instruction following.
Following the official protocol, we report the average score, which credits partial progress, and the success rate, each averaged over the tasks of a dimension and then over the five dimensions.
In view of the small items in the scenes, we use the head camera at $384\times480$ and the two wrist cameras at $192\times256$ in a T-shaped layout on the RoPE grid, and predict absolute end-effector poses in the robot base frame.
The episodes span 4 to 49 chunks, and we pack them with task-balanced ordering (\Cref{sec:infra_episode}) into 142 chunks per GPU.
We train with a peak learning rate of $1\times10^{-4}$ and a cosine schedule over 50k iterations with 32 GB300 GPUs.
}

\Cref{tab:sim_robodojo} compares \ourscausal with VLAs, WAMs and GPT-6 Astra, a frontier foundation model used directly as a policy.
With 4.5B parameters, \ourscausal achieves the best average among them, 34.31 / 27.09\%. Specifically, on the memory dimension, \ourscausal achieves a success rate of 52.33, outperforming GPT-6 Astra's 38.67 and InternW0-$\Delta$'s 34 by a large margin. In addition, \ourscausal achieves the best scores on the precision and long-horizon dimensions, demonstrating that our hybrid memory model supports both long-term history and precise control.

\subsection{RoboTwin 2.0}
\label{sec:sim_robotwin}

{The benchmark~\cite{chen2026robotwin2} contains 50 bimanual tasks on the ALOHA-AgileX robot with a head camera and two wrist cameras.
We follow the Clean2Random protocol: we post-train only on the 2,500 demonstrations collected in clean scenes, and evaluate the same checkpoint both in clean scenes and in randomized scenes with changed backgrounds, lighting, clutter and table height.
We use the same T-shaped camera layout and action space as on RoboDojo.
The episodes span 3 to 29 chunks, and each GPU packs whole episodes up to a budget of 15 chunks per step; a longer episode takes a GPU on its own and sets the budget of its step.
We train the model with a peak learning rate of $5\times10^{-5}$ and a cosine schedule over 20k iterations with 32 H200 GPUs.
}

\begin{wraptable}{R}{0.42\linewidth}
\centering
\caption{Results on RoboTwin 2.0~\cite{chen2026robotwin2} with clean-only training (Clean2Random).}
\label{tab:sim_robotwin}
\small
\begin{tabular}{lccc}
\toprule
Method & Clean & Rand. & Avg. \\
\midrule
\tablegroup{4}{VLA} \\
GR00T N1.7~\cite{nvidia2025gr00tn1} & 43.6 & 20.7 & 32.2 \\
StarVLA~\cite{starvla2026starvla} & 58.1 & 10.6 & 34.4 \\
X-VLA~\cite{zheng2025xvla} & 68.0 & 20.9 & 44.5 \\
Spatial Forcing~\cite{li2025spatialforcing} & 77.2 & 26.7 & 52.0 \\
ABot-M0~\cite{yang2026abotm0} & 70.7 & 36.0 & 53.4 \\
$\pi_{0.5}$~\cite{physicalintelligence2025pi05} & 73.1 & 47.9 & 60.5 \\
GigaBrain-0.7~\cite{gigabrain2026gigabrain07} & 66.8 & 67.9 & 67.4 \\
Qwen-RobotManip~\cite{yuan2026qwenrobotmanip} & 84.7 & 69.4 & 77.1 \\
\midrule
\tablegroup{4}{WAM} \\
AHA-WAM~\cite{cai2026ahawam} & 64.3 & 3.2 & 33.8 \\
Fast-WAM~\cite{yuan2026fastwam} & 77.8 & 1.9 & 39.9 \\
X-WAM~\cite{guo2026xwam} & 70.0 & 25.8 & 47.9 \\
4D-WAM~\cite{yang2026fourdwam} & 81.5 & 41.8 & 61.7 \\
OpenWAM-$\alpha$~\cite{wang2026openwam} & 89.4 & 48.7 & 69.0 \\
InternW0-$\Delta$~\cite{miao2026internw0delta} & 90.0 & {71.9} & 81.0 \\
\midrule
\textbf{\ourscausal} & \textbf{91.8} & \textbf{73.3} & \textbf{82.6} \\
\bottomrule
\end{tabular}
\end{wraptable}
\begin{table}[t]
\centering
\caption{Zero-shot success rate (\%) on LIBERO-Plus~\cite{fei2026liberoplus} of policies trained on the LIBERO~\cite{liu2023libero} dataset.}
\label{tab:sim_libero_plus}
\small
\setlength{\tabcolsep}{4pt}
\begin{tabular}{lcccccccc}
\toprule
Method & Camera & Robot & Language & Light & Background & Noise & Layout & Total \\
\midrule
\tablegroup{9}{VLA} \\
$\pi_0$~\cite{black2024pi0} & 13.8 & 6.0 & 58.8 & 85.0 & 81.4 & 79.0 & 68.9 & 53.6 \\
$\pi_0$-FAST~\cite{pertsch2025fast} & 65.1 & 21.6 & 61.0 & 73.2 & 73.2 & 74.4 & 68.8 & 61.6 \\
RIPT-VLA~\cite{tan2025riptvla} & 55.2 & 31.2 & 77.6 & 88.4 & 91.6 & 73.5 & 74.2 & 68.4 \\
OpenVLA-OFT~\cite{kim2025openvlaoft} & 56.4 & 31.9 & 79.5 & 88.7 & 93.3 & 75.8 & 74.2 & 69.6 \\
StarVLA~\cite{starvla2026starvla} & 52.5 & 49.8 & 88.5 & 95.7 & 95.7 & 73.0 & 76.9 & 74.1 \\
VLA-JEPA~\cite{sun2026vlajepa} & 63.3 & 67.1 & 85.4 & 95.6 & 93.6 & 66.3 & 85.1 & 79.5 \\
VLAct~\cite{yang2026vlact} & 73.9 & 68.4 & 81.5 & 96.7 & 96.7 & 86.0 & 83.3 & 82.6 \\
$\pi_{0.5}$~\cite{physicalintelligence2025pi05} & 78.4 & 73.6 & 80.8 & 96.2 & 94.1 & 89.0 & 84.5 & 84.4 \\
InternVLA-A1.5~\cite{ma2026internvlaa15} & 83.1 & 55.1 & 86.9 & 96.4 & 98.2 & 95.6 & 85.2 & 84.8 \\
Qwen-RobotManip-Context~\cite{yuan2026qwenrobotmanip} & 89.9 & 83.9 & 86.5 & 98.6 & \textbf{99.9} & \textbf{97.9} & 87.5 & 91.4 \\
\midrule
\tablegroup{9}{WAM} \\
Fast-WAM~\cite{yuan2026fastwam} & 16.4 & 44.5 & 68.9 & 78.2 & 53.7 & 37.7 & 60.7 & 51.5 \\
OpenWAM-$\alpha$~\cite{wang2026openwam} & 33.8 & 76.1 & 88.0 & 97.0 & 87.1 & 39.8 & 77.5 & 69.2 \\
4D-WAM~\cite{yang2026fourdwam} & 45.2 & 64.3 & 90.6 & 94.3 & 57.7 & 69.1 & 79.2 & 71.0 \\
ST-WAM~\cite{wang2026stwam} & 55.4 & 60.1 & 79.3 & 93.0 & 74.2 & 79.5 & 74.3 & 72.8 \\
Faster-WAM~\cite{ma2026fasterwam} & 67.9 & 49.0 & 92.1 & 94.3 & 57.0 & 82.3 & 82.7 & 75.0 \\
JEPA-WAM~\cite{lin2026jepawam} & 79.2 & 59.2 & 68.2 & 93.3 & 94.6 & 83.6 & 76.1 & 79.2 \\
Cosmos-Policy~\cite{kim2026cosmospolicy} & 75.8 & 63.3 & 81.7 & 96.5 & 88.9 & 92.7 & 82.2 & 82.2 \\
ImageWAM~\cite{zhang2026imagewam} & 80.8 & 50.3 & 91.4 & 98.1 & 85.5 & 93.8 & 80.5 & 83.1 \\
ABot-M0.5~\cite{chen2026abotm05} & 70.5 & 87.4 & 88.6 & 94.0 & 89.7 & 75.5 & 85.2 & 83.4 \\
Being-H0.7~\cite{luo2026beingh07} & 82.0 & 59.0 & 82.8 & 97.8 & 90.0 & 93.5 & 88.5 & 84.8 \\
InternW0-$\Delta$~\cite{miao2026internw0delta} & \textbf{90.6} & 91.1 & 92.9 & 95.8 & 99.4 & 94.0 & 88.4 & 92.8 \\
\midrule
\textbf{\ourscausal} & 84.1 & \textbf{98.4} & \textbf{93.9} & \textbf{99.7} & 98.8 & 94.4 & \textbf{91.8} & \textbf{94.0} \\
\bottomrule
\end{tabular}
\end{table}
On the RoboTwin 2.0~\cite{chen2026robotwin2} benchmark, \ourscausal achieves the best success rate on both clean and randomized settings. Such results demonstrate the effectiveness of our robot pretraining in improving the robustness of the world action model.

\subsection{LIBERO-Plus, LIBERO-Pro and LIBERO-Max}
\label{sec:sim_libero}

\begin{table}[t]
\centering
\begin{minipage}[t]{0.585\linewidth}
\centering
\captionsetup{justification=raggedright}
\caption{Zero-shot success rate (\%) on LIBERO-Pro~\cite{zhou2025liberopro} of policies trained on the LIBERO~\cite{liu2023libero} dataset.}
\label{tab:sim_libero_pro}
\small
\setlength{\tabcolsep}{4pt}
\begin{tabular}{lccccc}
\toprule
Method & Object & Position & Semantic & Task & Avg. \\
\midrule
\tablegroup{6}{VLA} \\
OpenVLA-OFT~\cite{kim2025openvlaoft} & 27.63 & 5.75 & 59.00 & 0.63 & 23.25 \\
NORA~\cite{hung2025nora} & 70.50 & 0.00 & 86.25 & 0.00 & 39.00 \\
MolmoAct~\cite{lee2025molmoact} & 76.00 & 1.50 & 85.75 & 1.50 & 41.00 \\
$\pi_0$~\cite{black2024pi0} & 90.50 & 0.00 & 90.50 & 0.00 & 45.25 \\
X-VLA~\cite{zheng2025xvla} & 79.00 & 0.75 & 96.75 & 6.75 & 46.00 \\
VLA-Adapter~\cite{wang2025vlaadapter} & 73.75 & 0.00 & 90.75 & 19.75 & 46.06 \\
OpenVLA~\cite{kim2024openvla} & 93.00 & 0.00 & 97.25 & 0.00 & 47.56 \\
SimVLA~\cite{luo2026simvla} & 81.50 & 8.30 & 98.80 & 6.00 & 48.60 \\
HABILIS Brain 0~\cite{pahk2026habilisbrain0} & 71.85 & 18.05 & 93.00 & 21.45 & 51.09 \\
$\pi_{0.5}$~\cite{physicalintelligence2025pi05} & \textbf{96.00} & 20.75 & 95.75 & 0.75 & 53.31 \\
Quantile Head~\cite{wang2026quantilehead} & 85.80 & 31.30 & 97.80 & 26.00 & 60.20 \\
QuoVLA~\cite{wang2026quovla} & 88.25 & 36.00 & \textbf{99.50} & 25.50 & 62.31 \\
Continuous Reasoning~\cite{wu2026continuousreasoning} & 84.60 & 39.30 & 94.90 & 37.10 & 64.00 \\
\midrule
\tablegroup{6}{WAM} \\
PatchWAM~\cite{wang2026patchwam} & 80.80 & 8.80 & 96.80 & 5.00 & 47.85 \\
Flex-$\pi$~\cite{yan2026flexpi} & 87.25 & 11.75 & 98.25 & 5.00 & 51.00 \\
GAM~\cite{zhang2026gam} & 75.75 & 32.75 & 95.00 & \textbf{40.50} & 61.00 \\
\midrule
\textbf{\ourscausal} & 88.95 & \textbf{47.65} & 98.10 & 37.65 & \textbf{68.09} \\
\bottomrule
\end{tabular}
\end{minipage}\hfill
\begin{minipage}[t]{0.385\linewidth}
\centering
\captionsetup{justification=raggedright}
\caption{Zero-shot success rate (\%) on LIBERO-Max~\cite{zhang2026liberomax} of policies trained on the LIBERO~\cite{liu2023libero} dataset.}
\label{tab:sim_libero_max}
\small
\setlength{\tabcolsep}{4pt}
\begin{tabular}{lcc}
\toprule
Method & Base & Dynamic \\
\midrule
\tablegroup{3}{VLA} \\
SmolVLA~\cite{shukor2025smolvla} & 26.1 & 15.0 \\
X-VLA~\cite{zheng2025xvla} & 62.6 & 37.7 \\
OpenVLA-OFT~\cite{kim2025openvlaoft} & 64.3 & 43.2 \\
GR00T N1.7~\cite{nvidia2025gr00tn1} & 69.3 & 50.0 \\
Xiaomi-Robotics-0~\cite{cai2026xiaomirobotics0} & 70.3 & 52.0 \\
VLA-JEPA~\cite{sun2026vlajepa} & 73.5 & 54.3 \\
DM0.5~\cite{dexmal2026dm05} & 79.8 & 62.1 \\
$\pi_{0.5}$~\cite{physicalintelligence2025pi05} & 79.7 & 65.7 \\
MolmoAct2~\cite{fang2026molmoact2} & 80.3 & 66.9 \\
\midrule
\tablegroup{3}{WAM} \\
Fast-WAM~\cite{yuan2026fastwam} & 42.0 & 24.0 \\
Light-WAM~\cite{li2026lightwam} & 54.8 & 37.3 \\
DiT4DiT~\cite{ma2026dit4dit} & 65.1 & 39.4 \\
HiMem-WAM~\cite{sun2026himemwam} & 73.0 & 57.7 \\
Cosmos-Policy~\cite{kim2026cosmospolicy} & 77.4 & 59.3 \\
\midrule
\textbf{\ourscausal} & \textbf{88.3} & \textbf{76.1} \\
\bottomrule
\end{tabular}
\end{minipage}
\end{table}

LIBERO~\cite{liu2023libero} is a single-arm benchmark with a Franka robot and four task suites: Spatial, Object, Goal and Long.
We post-train \ourscausal jointly on the demonstrations of the four suites and evaluate it on three robustness benchmarks built on them, including LIBERO-Plus~\cite{fei2026liberoplus}, LIBERO-Pro~\cite{zhou2025liberopro} and LIBERO-Max~\cite{zhang2026liberomax}.
For this post-training, we follow the pretraining image grid RoPE to place the external camera and wrist camera, each at a resolution of $256\times256$.
The episodes span 3 to 17 chunks; we train for 8k iterations with about 178 chunks per iteration, with a peak learning rate of $5\times10^{-5}$ and a cosine schedule over 20k iterations.

On LIBERO-Plus (\Cref{tab:sim_libero_plus}), \ourscausal succeeds in 94.0\% of all perturbed episodes, ahead of InternW0-$\Delta$ (92.8\%) and Qwen-RobotManip-Context (91.4\%).
On LIBERO-Pro (\Cref{tab:sim_libero_pro}), it achieves the best average, 68.1\% against 64.0\% for the next best method, largely through position perturbations, under which it succeeds in 47.7\% of the episodes, against at most 39.3\% for the other methods.
On LIBERO-Max (\Cref{tab:sim_libero_max}), it achieves the highest success rate in both the base (88.3\%) and the dynamic rollouts (76.1\%), outperforming the previous best dynamic model (MolmoAct2~\cite{fang2026molmoact2}) by nearly 10\%. These results demonstrate our foundation model has strong robustness and transfer ability.

\section{Real-World Experiments}
\label{sec:real}

\label{sec:real_tasks}

Beyond the benchmark, we design five tasks: three short tasks, Magic Cube, Three-Cup Shell Game and Air Hockey, and two long-horizon tasks, Make Coffee and Tower of Hanoi. We compare \ourscausal with baselines on Make Coffee, Tower of Hanoi and Magic Cube, and use Three-Cup Shell Game and Air Hockey to examine its memory and asynchronous control; on Tower of Hanoi and Air Hockey, we also compare inference with BF16, FP8 and NVFP4 GEMMs.

\subsection{Setup}

\noindent\textbf{Hardware and Training.}
Four tasks run on a YAM Ultra robot, and Make Coffee runs on a Galaxea R1.
For each task, we collect 100 episodes and train a separate policy on them, initialized from the pretrained \ourscausal (\S\ref{sec:method_policy}), for 10k steps on two H100 nodes; as in \S\ref{sec:infra_episode}, each step fills every GPU with a budget of 48 chunks rather than a fixed batch of episodes.
At deployment, the policy runs on an RTX PRO 6000 (Tower of Hanoi, Magic Cube, Three-Cup Shell Game and Air Hockey) or an RTX 5090 (Make Coffee).

\noindent\textbf{Evaluation.}
We evaluate every task over 20 rollouts, with the following metrics:
1) Three-Cup Shell Game: a rollout succeeds if the robot picks the cup that hides the ball;
2) Air Hockey and Magic Cube: each successful hit of the puck or turn of the cube counts as a success, and each failed one as a failure;
3) Tower of Hanoi: the number of rings stacked consecutively in the correct order, counted until the first misplaced ring and divided by the total of 7, averaged over the rollouts;
4) Make Coffee: the long-horizon task can be split into 7 sub-tasks, from picking up the capsule to handing over the coffee cup, and each rollout is scored by sub-tasks.

\subsection{Results}
\noindent\textbf{Long-Horizon and Precision.}
We compare \ourscausal with $\pi_{0.5}$~\cite{physicalintelligence2025pi05}, an open-source VLA, and OpenWAM-$\alpha$~\cite{wang2026openwam}, an open-source WAM, both fine-tuned on the same demonstrations of each task (\Cref{tab:real_baselines}).
Make Coffee and Tower of Hanoi test long-horizon execution: Make Coffee contains seven stages%
, and Tower of Hanoi requires stacking seven rings one after another; in both tasks, an error at one step compromises all the steps after it.
On Tower of Hanoi, \ourscausal stacks on average 83.6\% of the rings before the first error, against 45.7\% for OpenWAM-$\alpha$ and 44.3\% for $\pi_{0.5}$, and on Make Coffee it reaches 87.1\%, against 65.7\% and 57.1\%.
Tower of Hanoi and Magic Cube also require precise manipulation: each ring must be lowered with its hole aligned to the peg, and each turn must grasp one layer of the cube and rotate it without moving the others.
On Magic Cube, \ourscausal succeeds in 75\% of the turns, against 70\% for $\pi_{0.5}$ and 60\% for OpenWAM-$\alpha$.
These results agree with the long-horizon and precision dimensions of RoboDojo, on which \ourscausal also achieves the highest scores (\Cref{tab:sim_robodojo}).

\begin{table}[t]
\centering
\caption{Real-world comparison with baselines (\%). Make Coffee: the number of finished steps before the first error. Tower of Hanoi: rings stacked consecutively in the correct order before the first error, out of 7. Magic Cube: successful turns of the cube.}
\label{tab:real_baselines}
\small
\begin{tabular}{lccc}
\toprule
Method & Make Coffee & Tower of Hanoi & Magic Cube \\
\midrule
$\pi_{0.5}$~\cite{physicalintelligence2025pi05} & 57.1 & 44.3 & 70.0 \\
OpenWAM-$\alpha$~\cite{wang2026openwam} & 65.7 & 45.7 & 60.0 \\
\textbf{\ourscausal} & \textbf{87.1} & \textbf{83.6} & \textbf{75.0} \\
\bottomrule
\end{tabular}
\end{table}

\begin{figure}[t]
\centering
\begin{subfigure}[b]{0.235\linewidth}
\centering
\includegraphics[scale=1]{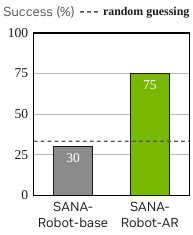}
\caption{Three-Cup Shell Game}
\end{subfigure}\hfill
\begin{subfigure}[b]{0.3895\linewidth}
\centering
\includegraphics[width=\linewidth]{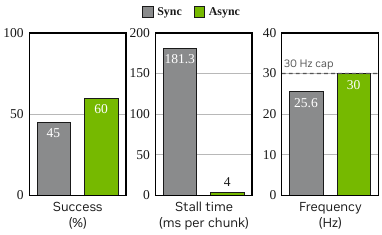}
\caption{Air Hockey}
\end{subfigure}\hfill
\begin{subfigure}[b]{0.3344\linewidth}
\centering
\includegraphics[width=\linewidth]{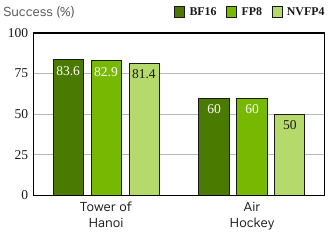}
\caption{BF16, FP8 and NVFP4}
\end{subfigure}
\caption{Real-world capabilities. (a) \oursbase and \ourscausal; the dashed line marks random guessing. (b) Synchronous and asynchronous execution of \ourscausal with BF16 GEMMs: success rate, stall time per chunk and effective control frequency; the dashed line marks the 30\,Hz cap of the controller. (c) Success on Tower of Hanoi and Air Hockey with BF16, FP8 and NVFP4 GEMMs.}
\label{fig:real_capabilities}
\end{figure}

\noindent\textbf{Long Memory.}
In the Three-Cup Shell Game, a ball is hidden under one of three cups and the cups are shuffled, after which the policy must lift the cup that hides the ball.
\ourscausal, with a window of $N=3$, succeeds in 75\% of the rollouts, whereas \oursbase, which sees only the current observation, succeeds in 30\%, no better than chance (\Cref{fig:real_capabilities}a).

\noindent\textbf{Fast Reaction.}
Air Hockey requires the robot to react to a fast-moving puck, so every pause between chunks costs hits.
In synchronous execution with BF16 GEMMs, the robot stalls for 181.3\,ms before each chunk while \ourscausal runs, and hits the puck in 45\% of the attempts.
Asynchronous execution with $d=8$ (\S\ref{sec:inf_pipeline}) overlaps inference with the execution of the current chunk and cuts the stall to about 4\,ms (network latency), raising the success rate to 60\% (\Cref{fig:real_capabilities}b).
It also keeps the effective control frequency at 30\,Hz, the cap of our controller, which executes each action for about 33\,ms, whereas the stalls of synchronous execution lower it to 25.6\,Hz.

\noindent\textbf{Low-Bit Inference.}
FP8 and NVFP4 GEMMs (\S\ref{sec:inf_opt}) reduce the latency of \ourscausal on RTX PRO 6000 by 20\% and 30\% over BF16, from 175.0\,ms to 140.6\,ms and 123.1\,ms (\Cref{tab:latency}), but quantize the weights and activations of the backbone to 8 and 4 bits.
We therefore deploy \ourscausal with BF16, FP8 and NVFP4 GEMMs on Tower of Hanoi and Air Hockey (\Cref{fig:real_capabilities}c).
Both low-bit variants largely retain the performance of BF16: FP8 reaches 82.9\% on Tower of Hanoi and 60\% on Air Hockey, against 83.6\% and 60\% with BF16, and NVFP4 reaches 81.4\% and 50\%, the latter amounting to two more missed hits in 20 attempts.
Low-bit GEMMs thus cut the latency by up to 30\% while keeping the success close to that of BF16.

\section{Memory Analysis}
\label{sec:memory}

In this section, we analyze the memory of \ourscausal along three questions.
\Cref{sec:mem_designs} compares it with the memory designs of VPP2 and InternW0-$\Delta$.
\Cref{sec:mem_obs} shows why the current observation must be a causal condition of the target chunk, rather than left out or attended bidirectionally.
\Cref{sec:mem_layers} examines what the softmax and the GDN layers each read from the memory.
Unless stated otherwise, we evaluate on the memory tasks of RoboDojo, which require information that is no longer visible in the current observation.
All variants in this section, including \ourscausal, are evaluated in our own runs rather than in the official evaluation, so the numbers of \ourscausal here differ from those in \Cref{tab:sim_robodojo}.

\subsection{Memory Designs}
\label{sec:mem_designs}

\begin{figure}[t]
\centering
\begin{subfigure}[b]{0.64\linewidth}
\centering
\includegraphics[width=\linewidth]{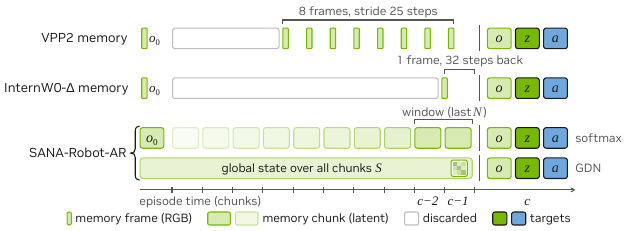}
\caption{History read by chunk $c$}
\end{subfigure}\hfill
\begin{subfigure}[b]{0.34\linewidth}
\centering
\includegraphics[width=\linewidth]{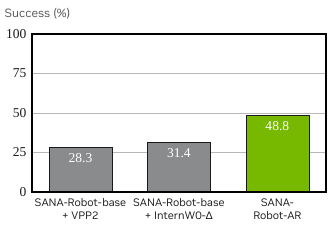}
\caption{Memory-task success}
\end{subfigure}
\caption{Memory designs on RoboDojo. The baseline memories are used to post-train \oursbase with the same data and initialization as \ourscausal; (b) reports the success rate on the memory tasks.}
\label{fig:mem_designs}
\end{figure}

We compare our memory with those of two recent WAMs by post-training \oursbase on RoboDojo with each of them, with the same data and initialization as \ourscausal (\Cref{fig:mem_designs}a):
1) VPP2 memory keeps the sink and 8 frames sampled backward from the current step at intervals of 1\,s;
2) InternW0-$\Delta$ memory keeps the sink and a single frame 32 steps back.
Both designs share three limitations.
First, their memory has a fixed capacity, so all history other than the kept frames is lost.
Second, sparsely sampled RGB frames are coarse, since they may miss the key motions or object states.
Third, the memory frames are encoded and forwarded again at every request, so the inference cost grows with the length of the memory.

\ourscausal instead uses a hybrid memory (\Cref{sec:method_memory}): the GDN layers keep a global state over all past chunks, and the softmax layers attend causally to the sink and the last $N$ chunks through a sliding window, which carries earlier history forward, chunk by chunk, into the current window.
The memory is chunk-level: each chunk stores the video latent of its full span of 32 control steps, so it keeps fine-grained information, such as every position the target cup passes through in the Three-Cup Shell Game (\Cref{sec:real_tasks}), which sparsely sampled frames may miss.
Each chunk is encoded and written into the memory once, after it has been observed (\Cref{sec:inf_pipeline}), so the memory covers the whole episode, and our model predicts each chunk at a constant cost per request.
This pays off in both performance and efficiency.
On the memory tasks of RoboDojo, \ourscausal reaches 48.8\% success, against 28.3\% and 31.4\% for \oursbase post-trained with the memory of VPP2 and InternW0-$\Delta$, respectively (\Cref{fig:mem_designs}b).
Our memory is particularly strong on tasks that need a long memory span: on press-by-number, where the policy must count how often it has pressed each button, \ourscausal succeeds in 82.67\% of the episodes in the official evaluation, against 3.3\% with the memory of InternW0-$\Delta$. 

\subsection{Conditioning on the Current Observation}
\label{sec:mem_obs}

\begin{figure}[t]
\centering
\begin{subfigure}[b]{0.17\linewidth}
\centering
\includegraphics[width=\linewidth]{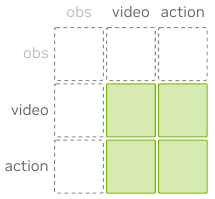}
\caption{w/o observation}
\end{subfigure}\hspace{0.8em}
\begin{subfigure}[b]{0.17\linewidth}
\centering
\includegraphics[width=\linewidth]{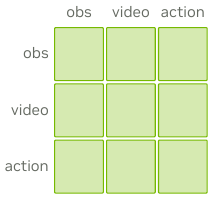}
\caption{Bidirectional}
\end{subfigure}\hspace{0.8em}
\begin{subfigure}[b]{0.17\linewidth}
\centering
\includegraphics[width=\linewidth]{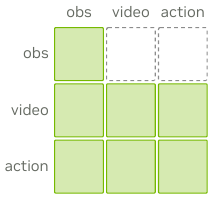}
\caption{\ourscausal}
\end{subfigure}\hspace{2em}
\begin{subfigure}[b]{0.37\linewidth}
\centering
\small
\setlength{\tabcolsep}{3.5pt}
\renewcommand{\arraystretch}{1.3}
\begin{tabular}{@{}lcc@{}}
\toprule
Variant & Memory & Precision \\
\midrule
w/o observation & 49.78 / 49.56 & 28.46 / 21.33 \\
Bidirectional & 10.38 / 9.11 & 35.47 / 26.08 \\
\midrule
\ourscausal & 49.10 / 48.78 & 39.41 / 30.08 \\
\bottomrule
\end{tabular}
\caption{RoboDojo results}
\end{subfigure}
\caption{Three ways to condition the target chunk on the current observation $\bm{o}_c$.
(a) $\bm{o}_c$ is not in the target chunk and appears only as the last frame of the previous chunk in the memory. (b) $\bm{o}_c$ and the targets attend to each other. (c) The targets read $\bm{o}_c$, but $\bm{o}_c$ does not read them and is not written into the memory. Filled: attention; dashed: no attention. (d) Average score / success rate (\%) on the memory and the precision tasks.}
\label{fig:mem_obs}
\end{figure}

The current observation is an essential input of a WAM.
In developing \ourscausal, we compared three ways for the target chunk to read it:
1) bidirectional: $\bm{o}_c$ and the targets attend to each other (\Cref{fig:mem_obs}b);
2) w/o observation: the target chunk has no observation token, and the current frame enters only through the memory, as the last frame of the previous chunk $\bm{v}_{c-1}$ (\Cref{fig:mem_obs}a);
3) causal (\ourscausal): the targets read $\bm{o}_c$, but $\bm{o}_c$ does not read them and is not written into the memory (\Cref{fig:mem_obs}c).
We evaluate them on the memory and the precision tasks of RoboDojo, which test the use of the history and of the current state, respectively (\Cref{fig:mem_obs}d).

We observe that each alternative fails at one of the two: the bidirectional policy ignores the memory, and the policy without the observation loses precision.
With bidirectional attention, the policy takes a shortcut through the current frame and succeeds in only 9.1\% of the memory tasks; it puts 14\% of its softmax attention on the memory (45\% without the observation), and removing the memory raises its action MSE by only $1.3\times$ ($249\times$ without the observation).
Without the observation, the policy is forced to use the memory (49.6\%), but it sees the current state only as part of the previous chunk, without the fine details of the current frame, so its precision drops to 21.3\%.
Our causal observation keeps both, with 48.8\% on the memory and 30.1\% on the precision tasks.
It also allows prefix caching: the keys and values of $\bm{o}_c$ are computed once per request and reused in all $K$ denoising steps (\Cref{sec:inf_opt}).

\subsection{How Softmax and GDN Layers Use Memory}
\label{sec:mem_layers}

\ourscausal is, to our knowledge, the first WAM with a hybrid memory, which raises a question that does not arise in WAMs with a single type of attention: what do the softmax and the GDN layers each learn to remember, and how does this history reach the actions?
We answer it offline, with teacher forcing on 140 RoboDojo episodes, using the intermediate checkpoints from 5k to 20k post-training steps.
To measure how much the policy relies on a source of history, we cut it from the forward pass while keeping the current observation, and report the action MSE relative to the full model, as a geometric mean over the chunks $c\ge2$, where $c$ is the index of a chunk within its episode (on the memory tasks unless stated otherwise).

\begin{figure}[t]
\centering
\begin{subfigure}[b]{0.49\linewidth}
\centering
\includegraphics[width=\linewidth]{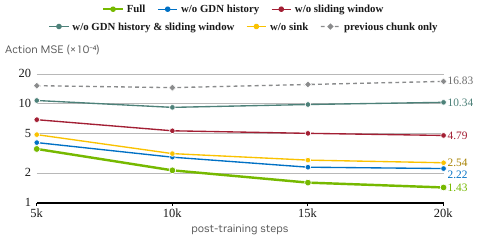}
\caption{Training dynamics}
\end{subfigure}\hfill
\begin{subfigure}[b]{0.49\linewidth}
\centering
\includegraphics[width=\linewidth]{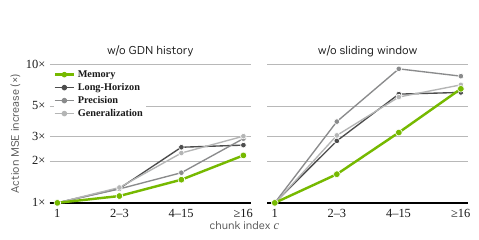}
\caption{Reliance by chunk index}
\end{subfigure}
\caption{
Memory analysis of \ourscausal 
offline on RoboDojo. (a) Action MSE on the memory tasks during post-training, with sources of history removed; ``w/o sliding window'' writes the past chunks without the window. (b) Increase of the action MSE at 20k steps when the GDN history is removed or the sliding window is cut, by chunk index $c$ and task category.
}
\label{fig:mem_layers}
\end{figure}

\noindent\textbf{Each Memory Component Matters.}
Removing any component of the memory sharply raises the action MSE at 20k steps (\Cref{fig:mem_layers}a): by $1.8\times$ without the sink ($5.3\times$ at $c=1$), $1.6\times$ without the GDN state, $3.4\times$ without the sliding window, $7.2\times$ without both, and $12\times$ with the previous chunk as the only history.
The policy learns to use these components during post-training: from 5k to 20k steps, the action MSE of the full model falls by $2.4\times$, while without the GDN state and the sliding window it stays flat, so the gains of post-training come from the memory.
The reliance holds in every category of RoboDojo, not only on the memory tasks (up to $19\times$ with the previous chunk alone).

\noindent\textbf{Softmax and GDN Layers Learn Memories of Different Granularity.}
The two types of layers carry the history beyond the previous chunk along two paths: the softmax layers pass exact latents forward chunk by chunk through the sliding window, while the GDN layers compress the whole episode into a fixed-size state.
The policy relies on both, and increasingly so as the episode grows (\Cref{fig:mem_layers}b).
On the memory tasks, cutting the sliding window raises the action MSE by $1.6\times$ for $c=2$--$3$ and by $6.7\times$ for $c\ge16$, and removing the GDN state raises it by $1.1\times$ and $2.2\times$, with the same trend in every category.
The coarse GDN state thus complements the fine-grained window over the long range.

\noindent\textbf{The Predicted Video Bridges the History to the Actions.}
In the softmax layers, the actions barely read the history directly: cutting these reads raises the action MSE by only $1.05\times$.
Instead, the video predicted for the current chunk reads the history, and the actions read this video: preventing the video from reading the history raises the action MSE by $3.8\times$, and cutting the read of the actions on the video raises it by $8.0\times$.
Predicting the video is therefore not only an auxiliary objective but a bridge through which the actions recall the past, which is also why the memory stores only video latents.

\FloatBarrier
\section{\ours World Simulator}
\label{sec:world_simulator}

Beyond acting as a policy, the FDM mode of \ours serves as a world simulator.
We first formulate the simulation problem (\S\ref{sec:method_simulator}), then describe how we post-train the simulator for long, few-step rollouts (\S\ref{sec:infra_post_training}), and finally evaluate it on WorldArena and on policy evaluation with DROID policies (\S\ref{sec:sim_experiments}).

\subsection{Problem Formulation}
\label{sec:method_simulator}

Our simulator starts from the FDM mode of the unified model.
For each chunk $c$, it takes the clean observation latent $\bm{o}_c$, robot state $\bm{s}_c$, and an action chunk $\bm{a}_c$ supplied by an external policy, and denoises the future video
latents $\bm{z}_c$.
Unlike the policy mode, which receives the real observations, the world simulator acts as the environment itself. Therefore, the continuous generation of the world simulator relies on its previous output as memory:
\begin{equation}
p\big(\bm{z}_{1:N} \mid \bm{\hat{o}}_1, \bm{s}_1, \bm{a}_{1:N}\big)
=
\prod_{c=1}^{N} p_\theta\big(\bm{z}_c \mid \bm{\hat{o}}_c, \bm{s}_c, \bm{a}_c, \bm{z}_{<c}\big),
\label{eq:simulator}
\end{equation}
where $\bm{\hat{o}}_c$ for $c>1$ is taken from the generated chunk $\bm{z}_{c-1}$, and $\bm{z}_{<c}$ is the history of generated chunks.
The actions $\bm{a}_c$ come from a policy that reads the generated observation.
Since each chunk conditions on previously generated rather than ground-truth content, errors can accumulate during the continuous rollout. Therefore, we introduce our world simulator post-training in this section to improve the long-range generation quality and efficiency.

\subsection{Long-Video Post-Training}
\label{sec:infra_post_training}

The world simulator is post-trained in two stages: (1) reducing the error accumulated with teacher forcing and resampling forcing; (2) improving the efficiency with few-step distillation.

\noindent\textbf{Stage 1: Teacher and Resampling Forcing.}
Starting from the pretrained \oursbase, we first adapt it to chunk-causal generation with teacher forcing, where every chunk conditions on the ground-truth history.
We then continue training with Resampling Forcing~\cite{guo2026resamplingforcing} to improve the robustness to rollout errors.
Specifically, we partially noise the ground-truth video latents and denoise them autoregressively with the current model, which yields a perturbed history.
Then the perturbed history is adopted as the condition for future chunks, enabling the model to tolerate errors in its history.

\noindent\textbf{Stage 2: Few-Step Distillation.}
We freeze the Stage-1 model as the teacher and distill a 4-step student from it with continuous-time consistency distillation (sCM)~\cite{lu2025scm}. 
The teacher provides the velocity at the sampled noise level, and the student learns to map every point on the teacher's probability-flow trajectory to the same clean endpoint.
After sCM, we apply Self-Forcing~\cite{huang2025selfforcing} to further reduce the accumulated error.

\subsection{Experiments}
\label{sec:sim_experiments}

A useful simulator should not only produce realistic video but also reproduce the task success that a policy reaches on the real robot.
We therefore evaluate it in two settings: action-conditioned video generation on WorldArena (\S\ref{sec:sim_worldarena}), and policy evaluation with DROID policies following Ctrl-World~\cite{guo2026ctrl} (\S\ref{sec:sim_droid}).

\subsubsection{WorldArena}
\label{sec:sim_worldarena}

\noindent\textbf{Benchmark and Protocol.}
WorldArena~\cite{shang2026worldarena} evaluates embodied world models on a curated subset of RoboTwin~2.0 (\S\ref{sec:sim_robotwin}): the Clean-50 configuration with 50 bimanual tasks and 50 episodes per task, 2,500 episodes in total for training and another 1,000 episodes for testing. 
The video-quality track leaderboard scores the generated videos with fifteen metrics grouped into six dimensions: visual quality, motion quality, content consistency, physics adherence, 3D accuracy and controllability.
EWMScore-P is the arithmetic mean of all metrics, scaled to 100.

\noindent\textbf{Results.}
\Cref{tab:worldarena} shows that \ours reaches an EWMScore-P of 66.8, second only to BWM. 
It obtains the highest visual quality and its weakest dimension is motion quality, which is likely due to the high temporal compression rate of our VAE. The 4-step student keeps most of the overall score and improves dramatically in terms of latency. 

\Needspace*{23\baselineskip}
\subsubsection{DROID Policy Evaluation}
\label{sec:sim_droid}

\begin{wrapfigure}{r}{0.37\linewidth}
\centering
\includegraphics[width=\linewidth]{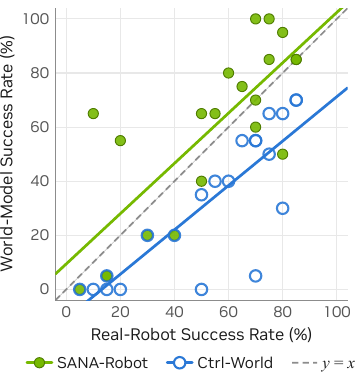}
\caption{Comparison of policy success rate in real robot and world simulator. The closer to the dashed line ($y=x$), the better the world simulator agrees with real deployment.
}
\label{fig:droid_success}
\end{wrapfigure}
\textbf{Benchmark and Protocol.}
Following Ctrl-World~\cite{guo2026ctrl}, we measure how well a world model can stand in for real-robot evaluation of generalist policies on the DROID platform~\cite{khazatsky2024droid}.
It evaluates three policies, $\pi_0$-DROID, $\pi_0$-FAST-DROID and $\pi_{0.5}$-DROID, on seven tasks: pick and place, towel folding, drawer, stacking, tissue, closing a laptop and table wiping, which gives 21 policy--task pairs with known real-robot success rates.
Each policy is rolled out in the world model on every pair, and human annotators label each rollout video as a success or a failure of the task; the annotators do not see which model generated a video, and all videos are re-encoded identically.
For a fair comparison, we also roll out the public Ctrl-World checkpoint under the same protocol, 20 rollouts per pair, and mix its videos with ours in one shuffled pool, so that both models are annotated blindly in the same round.

\noindent\textbf{Implementation Details.}
We evaluate a policy with policy-in-the-loop simulation: the policy reads the generated observation and proposes the next action chunk, and the simulator renders its visual consequences.
Repeating this process produces policy-in-the-loop rollouts entirely in imagination.
We generate the rollouts in the same way as Ctrl-World~\cite{guo2026ctrl}.
The simulator is trained on DROID with three camera views at $320\times192$ on the $2\times2$ grid, and is conditioned on absolute end-effector poses and joint positions.

\noindent\textbf{Results.}
\Cref{fig:droid_success} plots the world-model success rate of each pair against its real-robot success rate.
A faithful simulator puts the points on $y=x$.
Ctrl-World lies mostly below this line: its fit is $y=0.82x-10.8$, and its mean success rate over the 21 pairs is 32.4\%, against 52.6\% on the real robot.
The fit of \ours, $y=0.93x+9.4$, is closer to $y=x$ in slope, with a mean success rate of 58.3\%. 

\begin{table}[t]
\centering
\caption{Results on WorldArena~\cite{shang2026worldarena}. Latency is the end-to-end time in seconds to generate 30 frames at $640\times480$ on one H100 GPU.}
\label{tab:worldarena}
\small
\setlength{\tabcolsep}{4pt}
\begin{tabular}{lcccccccc}
\toprule
 & & \multicolumn{6}{c}{Dimension scores} & \\
\cmidrule(lr){3-8}
Method & EWMScore-P & Visual & Motion & Consist. & Physics & 3D Acc. & Control & Latency (s) \\
\midrule
TesserAct~\cite{zhen2025tesseract} & 54.9 & 38.9 & 38.9 & 65.9 & 39.5 & 79.1 & 77.7 & 95.1 \\
Cosmos-Predict 2.5~\cite{ali2025world} & 57.8 & 58.1 & 31.5 & 61.7 & 44.0 & 87.5 & 75.2 & 11.4 \\
IRASim~\cite{zhu2025irasim} & 59.6 & 56.1 & 32.5 & 65.7 & 49.3 & 88.8 & 77.4 & 11.9 \\
CogVideoX~\cite{yang2025cogvideox} & 62.7 & 55.8 & 42.2 & \textbf{67.7} & 50.9 & 88.3 & 82.6 & 58.7 \\
Ctrl-World~\cite{guo2026ctrl} & 63.7 & 57.4 & \textbf{50.9} & 62.2 & 55.4 & 88.5 & 78.2 & 38.8 \\
BWM~\cite{bwm2026bwm} & \textbf{68.6} & 63.5 & 42.0 & 67.6 & \textbf{69.0} & \textbf{97.3} & \textbf{88.6} & 4.9 \\
\midrule
\textbf{\ours} & 66.8 & \textbf{63.8} & 40.8 & 65.3 & 62.3 & 96.8 & 86.9 & 3.9 \\
\textbf{\ours (4-step)} & 66.2 & 63.1 & 39.2 & 64.9 & 62.8 & 96.5 & 86.6 & \textbf{0.3} \\
\bottomrule
\end{tabular}
\end{table}

\section{Related Work}
\label{sec:related_work}

\noindent{\textbf{Robot Policy Models.}
Vision-language-action models (VLAs) adapt pretrained vision-language models to output robot actions~\cite{brohan2023rt2,kim2024openvla,black2024pi0,physicalintelligence2025pi05,nvidia2025gr00tn1,zheng2025xvla}; they inherit the semantic knowledge of web-scale image-text data, but learn how a scene changes under their actions only from robot data.
A second line of work builds on video generation models, \textit{i.e.}, world action models (WAMs), which predict actions aligned with the future video~\cite{guo2024pad,li2025uva,zhu2025uwm,liao2025genie,kim2026cosmospolicy,ye2026dreamzero,li2026lingbotva,wang2026openwam,miao2026internw0delta,guo2026vpp2}.
Since WAMs inherit the cost of video generation, recent work makes them faster by decoding the actions without generating the video at test time~\cite{li2025uva,yuan2026fastwam,ni2026gigaworld,miao2026internw0delta}, by using a compact backbone or a lightweight action module~\cite{li2026lightwam,ma2026fasterwam}, by reducing the video to one or a few denoising steps~\cite{ye2026dreamzero,guo2026vpp2}, and by system-level optimization such as low-bit inference~\cite{ye2026dreamzero}.
\ours instead makes the backbone itself efficient: it builds on a hybrid linear video DiT~\cite{chen2026sana,zhao2026sana,zhu2026sana}, keeps generating the video, through which the actions read the history, and adds the deployment optimizations.}

\noindent{\textbf{Memory in World Action Models.}
Most VLAs and WAMs act from the current observation, although many manipulation tasks depend on what is no longer visible.
Therefore, history is crucial for some long-horizon tasks, but the cost of each decision grows with the length of the history.
VLAs thus compress the history: they keep a bank of perceptual and cognitive memory tokens~\cite{shi2026memoryvla}, aggregate features or compact tokens of past frames~\cite{li2025cronusvla,jang2025contextvla,koo2026hamlet}, or retrieve task-relevant keyframes~\cite{sridhar2025memer}.
Among WAMs, VPP2~\cite{guo2026vpp2} conditions on the first frame and eight past frames sampled at one-second intervals, and InternW0-$\Delta$~\cite{miao2026internw0delta} keeps the first frame and a single past frame 32 steps back; both encode these frames again at every request.
Autoregressive WAMs~\cite{ye2026dreamzero,li2026lingbotva, huang2026longwam} attend to a cache of past chunks with softmax attention, whose cost grows with the number of cached chunks.
\ours introduces a hybrid memory to WAMs, combining linear and softmax attention: the GDN state keeps a global memory of the whole episode at a constant cost, and the softmax window resolves the recent chunks in detail.}

\section{Conclusion}
\label{sec:conclusion}

{We presented \ours, a 4.5B world action model with a minute-level memory that runs in real time on edge devices.
Built on a hybrid linear diffusion transformer and pretrained on 19.4K hours of diverse data, \ours serves as a policy, a forward dynamics model and an inverse dynamics model.
Its memory pairs GDN layers, which keep the whole episode in a fixed-size state, with softmax attention over the first frame and the most recent chunks, so that every action chunk costs the same however long the episode runs.
Around this memory, we built the training and deployment infrastructure: multimodal sequence packing with task-balanced ordering trains on whole episodes without padding, memory RTC extends real-time chunking to a policy with memory, and fused kernels, CUDA graphs, prefix caching and low-bit GEMMs bring a request of \ourscausal to 115\,ms on an RTX 5090, which supports asynchronous control at 30\,Hz.
In addition to the efficient deployment, \ours achieves the best performance on several simulation benchmarks, outperforms $\pi_{0.5}$ and OpenWAM-$\alpha$ on real-world tasks and works as a simulator environment for DROID policies. In conclusion, \ours is an effective and efficient unified robotics foundation model.
}

{\footnotesize
\let\worlbibitem\bibitem
\renewcommand{\bibitem}{\par\worlbibitem}
\bibliographystyle{unsrt}
\bibliography{reference}
}

\end{document}